\documentclass[nohyperref]{article}

\usepackage{microtype}
\usepackage{graphicx}
\usepackage{subfigure}
\usepackage{booktabs} % for professional tables

\usepackage{hyperref}

\usepackage[accepted]{icml2022}

\usepackage{amsmath}
\usepackage{amssymb}
\usepackage{mathtools}
\usepackage{amsthm}
\usepackage{bbm}
\usepackage[capitalize,noabbrev]{cleveref}

\theoremstyle{plain}
\newtheorem{theorem}{Theorem}[section]

\newtheorem{lemma}[theorem]{Lemma}

\theoremstyle{definition}

\newtheorem{assumption}[theorem]{Assumption}
\theoremstyle{remark}

\usepackage[textsize=tiny]{todonotes}

\begin{document}

\twocolumn[
\icmltitle{Double Thompson Sampling in Finite stochastic Games}



\icmlsetsymbol{equal}{*}

\begin{icmlauthorlist}
\icmlauthor{Shuqing Shi}{yyy}
\icmlauthor{Xiaobin Wang}{comp}
\icmlauthor{Zhiyou Yang}{sch}
\icmlauthor{Fan Zhang}{sch}
\icmlauthor{Hong Qu}{equal,yyy}

\end{icmlauthorlist}

\icmlaffiliation{yyy}{Department of XXX, University of YYY, Location, Country}
\icmlaffiliation{comp}{Company Name, Location, Country}
\icmlaffiliation{sch}{School of ZZZ, Institute of WWW, Location, Country}

\icmlcorrespondingauthor{Firstname1 Lastname1}{first1.last1@xxx.edu}
\icmlcorrespondingauthor{Firstname2 Lastname2}{first2.last2@www.uk}

\icmlkeywords{Machine Learning, ICML}

\vskip 0.3in
]




\begin{abstract}
We consider the trade-off problem between exploration and exploitation under finite discounted Markov Decision Process, where the state transition matrix of the underlying environment stays unknown. We propose a double Thompson sampling reinforcement learning algorithm(DTS) to solve this kind of problem. This algorithm achieves a total regret bound of $\tilde{\mathcal{O}}(D\sqrt{SAT})$\footnote{The symbol $\tilde{\mathcal{O}}$ means $\mathcal{O}$ with log factors ignored} in time horizon $T$ with $S$ states, $A$ actions and diameter $D$. DTS consists of two parts, the first part is the traditional part where we apply the posterior sampling method on transition matrix based on prior distribution. In the second part, we employ a count-based posterior update method to balance between the local optimal action and the long-term optimal action in order to find the global optimal game value. We established a regret bound of $\tilde{\mathcal{O}}(\sqrt{T}/S^{2})$. Which is by far the best regret bound for finite discounted Markov Decision Process to our knowledge. Numerical results proves the efficiency and superiority of our approach.
\end{abstract}

\section{Introduction}
Online reinforcement learning~\cite{NIPS2017_36e729ec} mainly focus on the problem of learning and planning in sequential decision making systems in real time when the interacting environment is partially observed or fully observed. Normally, we could use Markov Decision Process(MDP) to represent such online decision process. At each time step, the system will generate reward and the next state according to a fixed state transition distribution. The decision maker tend to maximize the cumulative reward during its interacting process. Which leads to the trade-off between exploration and exploitation. Many attempts had been made to improve such dilemma~\cite{Kveton2020RandomizedEI}. In this paper, we aims to solve the trade-off problem in finite stochastic games between exploration and exploitation by applying posterior sampling method on policy probability distribution.

%exploring different states would help improve the accuracy of the model while exploiting the current information could increase the instantaneous return.

Trade-off between exploration and exploitation has been studied extensively in various scenarios such as stochastic games. The goal of exploration is to find as much information as possible of the environment. While the exploitation process aims to maximize the long-term reward based on the known environment. One of the popular way to deal with the trade-off problem is to use the Naive Exploration method such as adaptive $\epsilon$-greedy exploration~\cite{tokic2010adaptive}. It proposed a method that adjust the exploration parameter adaptively depend on the temporal-difference(TD) error observed form value function. Optimistic Initialisation methods have also been studied in factored MDP~\cite{inproceedings}~\cite{10.1162/153244303765208377} to solve trade-off problem. It encourages systematic exploration in early stage. Another common way to handle the exploitation-exploration trade-off is to use the optimism in the face of uncertainty (OFU) principle~\cite{lai1985asymptotically}. Based on this approach, the agent constructs confidence sets to search for optimistic parameters that associate with the maximum reward. Though Many of the optimistic algorithms were shown to have solid theoretical bounds of performance~\cite{auer2002using}~\cite{Hao2019BootstrappingUC}. They can still lock onto suboptimal action during exploration process.

Thompson Sampling(TS),also known as Posterior Sampling has been used in many scenes as an alternative strategy to promote exploration while balancing current reward. Thompson sampling was originally presented for stochastic bandit scenarios\cite{thompson1933likelihood}. Then it's been applied in various MDP contexts~\cite{osband2013more}. A TS algorithm estimate the posterior distribution of the unknown environment based on the prior distribution and experiment process. Theoretically, the TS algorithms tend to have tighter bounds than optimistic algorithms in many different contexts. Empirically, the TS algorithms can easily embedded with other algorithm structures because of its efficiency in computation~\cite{chapelle2011empirical}. The optimistic algorithms requires to solve all MDPs lying within the confident sets while TS algorithms only needs to solve the sampled MDPs to achieve similar results~\cite{russo2014learning}.

In this paper, we propose a sampling method that samples the transition probability distribution and policy distribution at the same time. Traditional posterior sampling method merely concentrate on the transition matrix of the underlying environment. Previous work \emph{UCSG} had given the regret upper bound of $\tilde{\mathcal{O}}\left(\sqrt[3]{D S^{2} A T^{2}}\right)$ on stochastic games~\cite{NIPS2017_36e729ec}. Where $D$ is the diameter of the Stochastic Games(SG). Model-free method has also been used in this area, \emph{Optimistic Q-Learning} achieves a regret bound of $\tilde{\mathcal{O}}\left(T^{2 / 3}\right)$ under infinite-horizon average discounted reward MDP~\cite{wei2020model}. Our approach consists of two optimization sampling method. The first method optimize the long-term policy probability distribution. The other method samples the transition matrix of the unknown environment. We first adopt the previous stopping criterions in Thompson Sampling-based reinforcement learning algorithm with dynamic
episodes (\emph{TSDE})~\cite{10.5555/3294771.3294898}. Then apply the posterior sampling method on both transition matrix and policy probability distribution. During the posterior update process of the policy distribution, we utilize the count-based update approach to represent the importance of each episode we sampled. Based on such approach, we managed to optimize the policy distribution in a time complexity of $\tilde{\mathcal{O}}(\sqrt{T}/S^{2})$ and transition probability distribution in $\tilde{\mathcal{O}}(D\sqrt{SAT})$.

\section{Preliminaries}

\subsection{Notations}
The finite stochastic game(FSG)~\cite{Cui2021MinimaxSC} could be defined by a 4-tuple $M=(\mathcal{S}, \mathcal{A}, r, \theta)$. Denote the size of the state space and the action space as $S=|\mathcal{S}|$ and $A=|\mathcal{A}|$. The reward function is defined as $r: S \times A \rightarrow \mathcal{R}$. And $\theta: S \times A \times S \rightarrow[0,1]$ represents the transition probability such that $\theta\left(s^{\prime} \mid s, a\right)=\mathbbm{P}\left(s_{t+1}=s^{\prime} \mid s_{t}=s, a_{t}=a\right)$. The actual transition probability $\theta_{*}$ is randomly generated before the game start. This probability is then fixed and unknown to agent. The transition probability in epoch $k$ and time step $t$ could be defined as $\theta_{t_{k}}$. After $T$ time step, the periodical transition probability could be represented as $\hat\theta_{k}$. A stationary policy $\pi: S \rightarrow A$ is a deterministic map that maps a state to an action. Therefore, we could define the instantaneous policy under transition probability $\theta_{t_{k}}$ as $\pi_{\theta_{t_{k}}}$. The local optimal policy under sub-optimal transition probability $\hat\theta_{k}$ could be represented as $\pi_{\hat\theta_{k}}$.
And the global optimal policy is defined as $\pi_{\theta_{*}}$ (The notation of the policy will be represented as $\pi_{\theta_{t_{k}}} = \pi_{t_{k}}, \pi_{\hat\theta_{k}} = \pi_{\hat{k}}, \pi_{\theta_{t_{k}}^{*}} = \pi_{t_{k}}^{*}, \pi_{\theta_{*}^{*}} = \pi_{*}$ for the sake of brevity).

In the FSG, the average discounted reward function per time step under stationary policy $\pi$ is defined as:
\begin{equation}
\label{equation1}
J_{\pi}(\theta)=\lim_{T \rightarrow \infty} \frac{1}{T} \mathbbm{E}\left[\gamma\sum_{t=1}^{T} r\left(s_{t}, a_{t}\right)\right]
\end{equation}
$\gamma$ is the discounted factor that satisfies $0 \textless \gamma \textless 1$. Therefore, we could denote the instantaneous average reward return under transition probability $\theta_{t_{k}}$ as $J_{\pi_{t_{k}}}(\theta_{t_{k}})$. Note that the $J_{\pi_{t_{k}}}(\theta_{t_{k}})$ is a theoretical value since its value is simulated under $\theta_{t_{k}}$, $\pi_{\theta_{t_{k}}}$. After $T$ step, the optimal average reward return $J_{\pi_{*}}(\theta_{t_{k}})$ could be deduced by the local optimal policy $\pi_{\hat\theta_{k}}^{*}$. The global optimal average reward return could be represented as $J_{\pi_{*}}(\theta_{*})$.

%\begin{equation}
%\label{equation2}
%J_{\pi_{t_{k}}}\left(\theta_{t_{k}}\right):=\lim_{T \rightarrow \infty} \frac{1}{T} E_{a_{t} \sim \pi_{\theta_{k t}}, s_{t} \sim \theta_{t_{k}}}\left[\sum_{t=1}^{T} r\left(S_{t}, a_{t}, \theta_{t_{k}}\right)\right]
%\end{equation}
%
%\begin{equation}
%\label{equation3}
%J_{\pi_{*}}\left(\theta_{t_{k}}\right):=\lim_{T \rightarrow \infty} \frac{1}{T} E_{a_{t} \sim \pi_{\theta_{k t}}^{*}, s_{t} \sim \theta_{t_{k}}}\left[\sum_{t=1}^{T} r\left(S_{t}, a_{t}, \theta_{t_{k}}\right)\right]
%\end{equation}
%
%\begin{equation}
%\label{equation4}
%J_{\pi_{*}}\left(\theta_{*}\right):=\lim_{T \rightarrow \infty} \frac{1}{T} E_{a_{t} \sim \pi_{\theta_{*}}^{*}, s_{t} \sim \theta_{*}}\left[\sum_{t=1}^{T} r\left(S_{t}, a_{t}, \theta_{*}\right)\right]
%\end{equation}

In the online learning setting, we use total regret to measure the performance of the decision maker. Total regret is defined as the difference between the total optimal game value and the actual game value.
\begin{equation}
Reg = \max_{a}\sum_{t=1}^{T}r(a,s_{t})-\sum_{t=1}^{T}r(a_{t},s_{t})
\end{equation}

Normally, such metric could be hard to calculate. Therefore, we define the \emph{bias vector} $b(\theta,\pi,s)$ as the relative advantage of each states to help us measure the total regret.
\begin{equation}
b(\theta, \pi, s):=E\left[\sum_{t=1}^{\infty} r\left(s_{t}, a_{t}\right)-J(\theta) \mid s_{1}=s, a_{t} \sim \pi(\cdot |s_{t})]\right.
\end{equation}
Under stationary policy $\pi$, the advantage between state $s$ and $s^{\prime}$ is defined as the difference between the accumulated reward with initial state $s$ and $s^{\prime}$.Which will eventually converge to the difference of its bias vectors $b(\theta,\pi,s)-b(\theta,\pi,s^{\prime})$. The bias vector satisfies the Bellman equation. Out of brevity, we denote the the expected total reward under stationary policy $\pi$ as $r(s,\pi) = E_{a \sim \pi(\cdot|s)}[\sum r(s,a)]$. The expected transition probability is denoted as $p_{\theta}\left(s^{\prime} \mid s, \pi\right) = E_{a \sim \pi(\cdot|s)}[p_{\theta}(s^{\prime}|s,a)]$. The Bellman equation is shown in Equation \ref{equation5}.
\begin{equation}
\label{equation5}
J(\theta, \pi, s)+b(\theta, \pi, s)=r(s, \pi)+\sum_{s^{\prime}} p_{\theta}\left(s^{\prime} \mid s, \pi\right) b(\theta, \pi, s^{\prime})
\end{equation}
In order to represent the difference between each state, we define \emph{span(h)} as $sp(b) = max(b) - min(b)$. The regret is strongly connected to $sp(b(\theta_{*},\pi_{\theta_{*}^{*}},\cdot)$. And for any $b(\theta,\pi,\cdot)$, we have $sp(b(\theta,\pi,\cdot) \leq \max_{s,s^{\prime}}T_{s \rightarrow s^{\prime}}^{\pi}(\theta) = D$. This represents the span of vector $b$ is less than or equal to the maximum expected time to reach to state $s^{\prime}$ from state $s$ under transition probability $\theta$ and policy $\pi$.

\subsection{Problem Setting}
When dealing with the non-convex SGs, the global optimal policy may be hard to get.  Because they sometimes stuck in local optimal results. The $\epsilon$ tolerance is then introduced to help measure the ability of the algorithm. When the difference between the optimal average return and the current average return is less than constant $\epsilon$. We could consider the current policy as the $\epsilon$-optimal policy.
\begin{assumption}
\label{assumption1}
\textbf{($\epsilon$-Optimal policy)}Under suboptimal and optimal transition probability, if policy $\pi_{t_{k}}$,$\pi_{\hat{k}}$ satisfies
$$
J_{\pi_{*}}(\theta_{t_{k}}) - J_{\pi_{t_{k}}}(\theta_{t_{k}}) \leq \epsilon
$$
$$
J_{\pi_{*}}(\theta_{*}) - J_{\pi_{\hat{k}}}(\theta_{*}) \leq \epsilon
$$
Then, policy $\pi_{t_{k}}$,$\pi_{\hat{k}}$ is $\epsilon$-optimal.
\end{assumption}

Assumption \ref{assumption2} implies that under all circumstances, all the states could be visited in average $D$ steps. When the agent conduct optimal policy under the optimal transition probability, the transition time $T_{s \rightarrow s^{\prime}}^{\pi_{*}}(\theta_{*})$ should be the shortest. Because the agent tend to explore the fewest non-related state with the optimal stationary policy. In a similar fashion, the transition time $T_{s \rightarrow s^{\prime}}^{\pi_{t_{k}}^{*}}(\theta_{t_{k}})$ for agent which conducts optimal policy under suboptimal transition probability should be less than the transition time $T_{s \rightarrow s^{\prime}}^{\pi_{t_{k}}}(\theta_{t_{k}})$ in the normal settings.

\begin{assumption}
\label{assumption2}
\textbf{(Expected transition time)}When conducting stationary policy $\pi$, assume the maximum expected time to reach to state $s^{\prime}$ from state $s$ under suboptimal transition probability and optimal transition probability is less than constant $D$:
$$
\max T_{s \rightarrow s^{\prime}}^{\pi_{*}}(\theta_{*}) \leq \max T_{s \rightarrow s^{\prime}}^{\pi_{t_{k}}^{*}}(\theta_{t_{k}}) \leq  \max T_{s \rightarrow s^{\prime}}^{\pi_{t_{k}}}(\theta_{t_{k}}) \leq D
$$
\end{assumption}

Let $e(t):=k$ be the epoch where the time instant $t$ belongs. Define $\mathcal{H}_{s_{1},s_{2}}(k,\pi)$ as the set of all the time instants that the state transition $s_{1} \rightarrow s_{2}$ occurs in the first $k$ epochs when stationary policy $\pi$ was used.
\begin{equation}
\begin{aligned}
&\mathcal{H}_{\left(s_{1}, s_{2}\right)}(k, \pi) \\
&:=\sum_{t=1}^{\infty} \mathbbm{1}\left\{\pi_{e(t)}=\pi,\left(S_{t}, S_{t+1}\right)=\left(s_{1}, s_{2}\right), N(e(t)) \leq k\right\}
\end{aligned}
\end{equation}
Under transition probability $\theta_{t_{k}}$, the expected transition time from state $s$ to state $s^{\prime}$ with stationary policy $\pi_{t_{k}}$ could be denoted as $\tilde\tau_{\pi_{t_{k}}}$, which satisfies $\tilde\tau_{\pi_{t_{k}}} = T_{s \rightarrow s}^{\pi_{t_{k}}}(\theta_{t_{k}})$. Therefore, the posterior probability of the stationary policy $\pi$ could be represented as the difference between the empirical state pair frequency $\frac{\mathcal{H}_{\left(s_{1}, s_{2}\right)}\left(k, \pi\right)}{k}$ and the corresponding expected value $\tilde\tau_{\pi_{t_{k}}}$.
\begin{assumption}
\label{assumption3}
\textbf{(Posterior distribution under suboptimal trajectories)}For any given scalars $e_1,e_2 \geq 0$, there exists $p \equiv p(e_1,e_2) \textgreater 0$ satisfies $\theta_{t_{k}}(\pi_{t_{k}}^{*}) \geq p$ for any epoch index $k$ at which suboptimal transition frequencies have been observed:
$$
\begin{aligned}
&\left|\frac{\mathcal{H}_{\left(s_{1}, s_{2}\right)}\left(k, \pi\right)}{k} - \tilde{\tau}_{\pi_{t_{k}}}\theta \left(s_{1}|s_{2}\right)\right| \leq \sqrt{\frac{e_{1} \log \left(e_{2} \log k\right)}{k}} \\
&\forall s_{1}, s_{2} \in \mathcal{S}, k \geq 1, c \in \mathcal{C}, k=\sum_{\pi \in \Pi} k
\end{aligned}
$$
\end{assumption}

Under finite discounted Markov decision processes, the average discounted return is also finite. So, we define the maximum average discounted reward as $\Gamma$. Which is the maximum reward that an agent could achieve during its exploration in the finite discounted Markov decision processes. The maximum value will be achieved under optimal transition probability with optimal stationary policy.

\begin{assumption}
\label{assumption4}
\textbf{(Upper bound for the average discounted reward)}Under the finite discounted MDP, the maximum average discounted reward is bounded by a constant.
$$
J_{\pi_{*}}(\theta_{*}) \leq \Gamma
$$
\end{assumption}

Based on the upper assumptions, we could then construct our method.

\section{Method}

In this section, we propose the Double Thompson Sampling method. One of the essential parameters under Thompson Sampling setting is the prior distribution. Which is denoted as $\mu_{0}$ in our paper. Note that we generate prior distribution for both transition probability and stationary policy. In each epoch $k$, at each time step $t$, the posterior distribution $\mu_{t_{k}}$ will be updated based on the previous history $h_{t_{k}}$. Let $N_{t}(s,a)$ be the number of visits to any state-action pair $(s,a)$ during a period of time $t$.
\begin{equation}
N_{t}(s, a)=\left|\left\{\tau<t:\left(s_{\tau}, a_{\tau}\right)=(s, a)\right\}\right|
\end{equation}
Therefore, the algorithm could be generated as follows.
\begin{algorithm}[h]
\caption{Double Thompson Sampling}
\label{alg1}
\textbf{Input}: Game Environment, Prior Distribution for transition probability $\mu_{\theta_{0}}$, Prior Distribution for stationary policy $\mu_{\pi_{0}}$, Transition Probability $\theta_{0}$, Initial State $s_{0} \in S$\\
\textbf{Output}: Stationary Policy $\pi_{K}$
\begin{algorithmic} %[1] enables line numbers
\FOR{Episode $k = 0,1,2 \dots K$}
\STATE $T_{k-1} \leftarrow t-t_{k}$
\STATE $t_{k} \leftarrow t$
\STATE Generate $\mu_{k}(\hat\theta_{k})$, $\mu_{k}(\hat \pi_{k})$ based on prior distribution
\FOR {$t \leq t_{k}+T_{k-1}$ and $N_{t}(s,a) \leq 2N_{t_{k}}(s,a)$}
\STATE Apply action $a_{t} \sim \pi_{t_{k}}, \pi_{t_{k}} \sim \mu_{t_{k}}(\pi)$
\STATE Observe new state $s_{t+1}$, reward $r_{t+1}$
\STATE Update posterior distribution $\mu_{t+1_{k}}(\pi), \mu_{t+1_{k}}(\theta)$ using EVI
\STATE $t \leftarrow t+1$
\ENDFOR
\ENDFOR
\end{algorithmic}
\end{algorithm}

The Double Thompson Sampling method(Alg \ref{alg1}) is conducted in multiple steps. At the beginning of each epoch $k$, the algorithm estimates the periodical transition probability using the past history $\mu_{k-1}(\hat\theta_{k-1})$(Step 1). This prior distribution satisfies $\mu_{k-1}(\hat\theta_{k-1}) = \mu_{(k-1)_{T}}(\theta_{(k-1)_{T}})$. We set two stopping criterion for our algorithm in order to limit our agent's exploration direction. The first stopping criterion aims to stop meaningless exploration. The second stopping criterion ensures that any state-action pair $(s,a)$ will not be encounter twice during the same epoch.  During each epoch $k$, actions are generated from the instantaneous policy $\pi_{t_{k}}$(Step 3). This policy follows a posterior distribution $\mu_{t_{k}}(\pi)$. These actions are then be used by the agent to interact with the environment to observe the next state $s_{t+1}$ and the reward $r_{t+1}$(Step 4). The observation results are then be used to find the optimal posterior distribution for policy $\pi_{t+1_{k}}$ and transition probability $\theta_{t+1_{k}}$(Step 5). If the stopping criterions are not met, the algorithm will start over from Step 2. The whole process will be repeated until the terms of the stopping criterions are met.

\subsection{Update Rule}
In the model-based method, the update method of the transition probability is of great importance. Our method is a Thompson Sampling-based method. The transition probability will be updated based on the prior distribution. Based on the Bayes' rule, the posterior distribution of the transition probability could be represented as :
\begin{equation}
\mu_{t+1}(\theta)=\frac{\theta\left(s_{t+1} \mid s_{t}, a_{t}\right) \mu_{t_{k}}(\theta)}{\sum_{\theta^{\prime} \in \Theta} \theta^{\prime}\left(s_{t+1} \mid s_{t}, a_{t}\right) \mu_{t}\left(\theta^{\prime}\right)}
\end{equation}

The update method of the stationary policy is different from the one of transition probability. In this paper, we introduce the prior policy to guide the current policy. Using the Thompson sampling and the Policy Iteration method(EVI), the algorithm is able to balance between the current optimal action and the history optimal action. This will help our method achieve long-term maximum return. Which is the global optimal value in this scenario. Let $W_{t_{k}}$ be the \emph{posterior weight} in epoch $k$ at time $t$. $J_{\pi_{t}}(\theta)$ and $J_{\pi^{*}}(\theta)$ denotes the instantaneous average discounted return and the local optimal value.
\begin{equation}
W_{t_{k}}(\pi) = \exp \sum_{\pi,s}\mathcal{H}(N_{\pi}(k),\pi)\log \frac{J_{\pi_{t}}(\theta)}{J_{\pi^{*}}(s,\theta)}
\end{equation}

Its value is  proportional to the log difference between the average return of the local optimal policy and current policy. Using the posterior factor, we could generate the Policy Iteration method based not only on the current observation but also the historical trajectory.
\begin{algorithm}[h]
\caption{Policy Iteration with Posterior factor}
\label{alg2}
\textbf{Input}: Game Environment, Prior Distribution for stationary policy $\mu_{t}(\pi)$, $0 \textless \gamma \textless 1$\\
\textbf{Output}: Stationary Policy $\pi_{i}$
\begin{algorithmic} %[1] enables line numbers
\REPEAT
\STATE $\mu_{t}(\pi) = W_{t}  \mu_{t-1}(\pi) + (1-W_{t})\pi_{t}^{*}(s,\theta_{t_{k}})$
\UNTIL {$D_{\theta}(\mu_{*}(\pi)||\mu_{t_{k}}(\pi)) \leq \epsilon$}
\end{algorithmic}
\end{algorithm}

The posterior distribution $\mu(\pi)$ is defined as the transition matrix under time step $t$. Satisfying $\mu_{t}(\pi) = \left(\mathcal{H}_{s_1,s_2}(t,\pi)\right)_{s_1,s_2 \in S}$. In this paper, we could denote the distance between the history optimal policy and the instantaneous policy using the \emph{Marginal Kullback-Leibler Divergence}(Marginal KL Divergence). Marginal KL Divergence is a widely used metric when measuring the difference between two probability distribution. Therefore, the distance could be represented as $D_{\theta}(\mu_{*}(\pi)||\mu_{t_{k}}(\pi))$.
\begin{equation}
\begin{aligned}
D_{\theta}(\mu_{*}(\pi)||\mu_{t_{k}}(\pi)) &:= \sum_{s_{1} \in \mathcal{S}} \theta_{s_{1}}^{\pi} \sum_{s_{2} \in \mathcal{S}} \mu_{*}(\pi) \log \frac{\mu_{*}(\pi)}{\mu_{t_{k}}(\pi)} \\
&=\sum_{s_{1} \in \mathcal{S}} \theta_{s_{1}}^{\pi} \mathbbm{K} \mathbbm{L}\left(\mu_{*}(\pi) \| \mu_{t_{k}}(\pi)\right)
\end{aligned}
\end{equation}
The marginal KL divergence $D_{\pi}(\mu_{*}(\theta)||\mu_{t_{k}}(\theta))$ is a convex combination between the history optimal policy and the instantaneous  policy. Parameter $\epsilon$ represents the tolerance between the optimal policy and the instantaneous policy. This posterior policy iteration(PPI) method updates the policy dynamically with the posterior factor. The policy will converge to optimal value after certain amount of iterations under this update method. In the following section, we will be introducing the proof of the astringency of this posterior update method.

\section{Main Results}
\subsection{Astringency of the Update Rule}
\label{section4.1}
In the online learning domain, one of the basic metric of an algorithm is whether it could converge after constant number of steps. So in this section, we provide the proof of the astringency of our posterior update method in order to illustrate the superiority of our method.

The following three Lemmas are meant to prove the convergence of our algorithm. In Lemma \ref{lemma1}, We first prove that for stochastic games $M$, the PPI method converges asymptotically. Then, in Lemma \ref{lemma0}, we demonstrate that the output policy of such policy iteration method updates monotonically towards optimal direction. Which is a vital evidence for the global optimality of our update method. At last, the third lemma(Lemma \ref{lemma10}) proves that under stochastic games $M$, the output policy generated from PPI method would reach $\epsilon$-optimal after constant number of iterations. 
\begin{lemma}
\label{lemma1}
Suppose Assumption \ref{assumption2} holds for some stochastic games $M$, then the policy iteration algorithm on $M$ converges asymptotically.

\begin{proof}
If the Assumption \ref{assumption2} holds. From Theorem 4 in ~\cite{72bf2cf9895a47e7be9a668a25215110}, the policy iteration converges.
\end{proof}
\end{lemma}

\begin{lemma}
\label{lemma0}
Under update algorithm PI, the average discounted return should be monotonically increased.
\begin{proof}
From Algorithm \ref{alg2}, we could deduce the update rule of the average discounted return:
\begin{equation}
\begin{aligned}
J_{\pi_{t}}(\theta) - J_{\pi_{t-1}}(\theta) &= (W_{t}-1)J_{t-1}(\theta) + (1-W_{t})J_{\pi^{*}}(s,\theta)\\
&= (1-W_{t})(J_{\pi^{*}}(s,\theta)-J_{\pi_{t-1}}(\theta))
\end{aligned}
\end{equation}
When $J_{\pi^{*}}(s,\theta) \geq J_{\pi_{t-1}}(\theta)$, we could deduce that $\log\frac{J_{\pi_{t}}(\theta)}{J_{\pi^{*}}(s,\theta)} \leq 1$. So the posterior weight $W_{t}$ is less than 1. This result holds vice versa. The first term $1-W_{t} \leq 0$ when $J_{\pi^{*}}(s,\theta) \leq J_{\pi_{t-1}}(\theta)$. Therefore, we could prove that:
\begin{equation}
J_{\pi_{t}}(\theta) - J_{\pi_{t-1}}(\theta) = (1-W_{t})(J_{\pi^{*}}(s,\theta)-J_{\pi_{t-1}}(\theta)) \geq 0
\end{equation}
The sequence $J_{\pi_{t}}(\theta)$ is monotonically increased with time step $t$.
\end{proof}
\end{lemma}

\begin{lemma}
\label{lemma10}
Suppose Assumption \ref{assumption1} and Assumption \ref{assumption2} hold for some stochastic games $M$. Let $v_{i}$ be the state value in iteration $i$. Define $N$ as the maximum iteration number of the algorithm. Then $\pi_{t_{k}}$ is $\epsilon$-optimal after $N$ iterations.
\begin{proof}
Define $D = \min_{s}\{\mu_{i+1}(\pi)-\mu_{\pi}\}$ and $U = \max_{s}\{\mu_{i+1}(\pi)-\mu_{i}(\pi)\}$. Then we could deduce:
\begin{equation}
\begin{aligned}
D + \mu_{N}(\pi) &\leq \mu_{N+1}\\
&\leq W_{i}\mu_{N} + (1-W_{i})\pi_{i}^{*}(s,\theta)\\
&\leq W_{i}\mu_{N} + (1-W_{i})(r_{N} + \theta_{} v_{N})
\end{aligned}
\end{equation}
Since $0 \textless W_{i} \textless 1$, the upper equation could be turned to:
\begin{equation}
D \leq (1-W_{i})J_{\pi_{i}}(\theta)
\end{equation}
Let $\pi^{*}$ be the optimal policy under all states that satisfies $\pi^{*} := \sum_{s \in S}\pi_{i}^{*}(s,\theta)$. Then
\begin{equation}
D \leq (1-W_{i})J_{\pi_{i}}(\theta) \leq (1-W_{i})J_{\pi^{*}}(\theta)
\end{equation}
In a similar way, we could also prove $U \geq (1-W_{i})J_{\pi^{*}}(\theta)$. From the definition of the stopping criterion of the Policy Iteration algorithm, we could assume $U - D \leq (1-W_{i})\epsilon$. Therefore, we have
\begin{equation}
\begin{aligned}
U &\leq D + (1-W_{i})\gamma \\
%U &\leq (1-W_{i})J_{\pi_{i}}(\theta)+(1-W_{i})\epsilon\\
U &\leq (1-W_{i})(J_{\pi_{i}}(\theta)+\epsilon)\\
(1-W_{i})J_{\pi^{*}} &\leq (1-W_{i})(J_{\pi_{i}}(\theta)+\epsilon)\\
J_{\pi^{*}} &\leq J_{\pi_{i}}(\theta)+\epsilon
\end{aligned}
\end{equation}
We could deduce that stationary policy $\pi$ is $\epsilon$-optimal after $N$ iterations.
\end{proof}
\end{lemma}

\subsection{Regret Bound Analysis}

After proving the astringency of the PPI method. We then move the proof of the regret bound. Which is the most popular metric for online learning method. Inside each episode, the regret could be separated into three parts. We could know the regret in time step $T$ would be represented as:
\begin{equation}
\begin{aligned}
Reg_{T} &= TJ_{\pi_{\hat{k}}}(\hat \theta) - \sum_{t = 1}^{T}r_{\pi_{t}}(s_{t},a_{t})\\
&= Reg_{T}^{1} + Reg_{T}^{2} + Reg_{T}^{3}
\end{aligned}
\end{equation}
We could define the following regret as:
\begin{equation}
\label{equation17}
\begin{array}{l}
Reg_{T}^{1} = TJ_{\pi_{\hat{k}}}(\hat \theta) - \sum_{t = 1}^{T}J_{\pi_{t}}(\hat \theta) \\ \\
Reg_{T}^{2}=\sum_{t = 1}^{T}J_{\pi_{t}}(\hat \theta) - \sum_{t = 1}^{T}J_{\pi_{t}}(\theta_{t}) \\ \\
Reg_{T}^{3}= \sum_{t = 1}^{T}J_{\pi_{t}}(\theta_{t}) - \sum_{t = 1}^{T}r_{\pi_{t}}(s_{t},a_{t})
\end{array}
\end{equation}
Where $J_{\pi_{\hat{k}}}(\hat{\theta})$ is the terminal average reward under terminal policy $\pi_{\hat{k}}$ and transition probability $\hat{\theta}$. Note that this value is a virtual value and only exists in theoretical analysis. $Reg_{T}^{1}$ represents the posterior difference between the total episodic reward and the total virtual instantaneous reward. We could assume such difference is calculated under the same transition probability since the transition probability is generated from the same priors. Since for any measurable function f and any $h_{t_{k}}$-measurable random variable $X$,$\mathbb{E}\left[f\left(\theta_{*}, X\right) \mid h_{t_{k}}\right]=\mathbb{E}\left[f\left(\theta_{k}, X\right) \mid h_{t_{k}}\right]$. This has been proved in previous studies.~\cite{osband2013more}

In order to bound the terminal regret $Reg_{T}^{1}$, we first bound the ratio between the expected optimal average discounted reward and the instantaneous discounted reward. Based on Assumption \ref{assumption1} and Assumption \ref{assumption4}, the expected optimal reward that an agent could achieve in the finite discounted MDP could be bounded by parameter $\Gamma$ and $\epsilon$.

%\begin{theorem}
%The regret in time step $T$ could be bounded by:
%$$
%Reg_{T} \leq \tilde{\mathcal{O}}(\mathcal{D\sqrt{SAT}})
%$$
%\end{theorem}

\begin{lemma}
$$
\log \frac{J_{\pi_{*}}(\theta)}{J_{\pi_{t}}(\theta)} \leq \frac{\epsilon}{\Gamma}
$$
\begin{proof}
First, we could multiply $J_{\pi_{t}}(\theta)$ in order to construct the inequality. Let $J_{\pi_{t}}(\theta) = n$, $\epsilon = x$
\begin{equation}
\begin{aligned}
\lim _{n \rightarrow+\infty}\left(1+\frac{x}{n}\right)^{n} &= \lim _{n \rightarrow+\infty} e^{n \ln \left(1+\frac{x}{n}\right)} \\
&=e^{\lim _{n \rightarrow+\infty} \frac{\ln \left(1+\frac{x}{n}\right)}{\frac{1}{n}}}
\end{aligned}
\end{equation}
Apply the L'Hopital's Rule:
\begin{equation}
\begin{aligned}
\lim _{n \rightarrow+\infty}\left(1+\frac{x}{n}\right)^{n}&= e^{\lim _{n \rightarrow+\infty} \frac{\left(\frac{-x}{n^{2}}\right) \frac{1}{1+\frac{x}{n}}}{-\frac{1}{n^{2}}}}\\
&= e^{\lim _{n \rightarrow+\infty} \frac{x}{1+\frac{x}{n}}}=e^{x}
\end{aligned}
\end{equation}
Then, we could prove that $\left(1+\frac{x}{n}\right)^{n}$ is monotonically increased with $n$:
\begin{equation}
\begin{aligned}
(1+\frac{x}{n})^{2} &= 1 \cdot\underbrace{ \left(1+\frac{x}{n}\right) \cdot\left(1+\frac{x}{n}\right) \cdots \cdots\left(1+\frac{x}{n}\right)}_{n}\\
& \leq \left[\frac{1+(1+\frac{x}{n})+ \cdots +(1+\frac{x}{n})}{n+1}\right]^{n+1}\\
& = \left[\frac{1+n(1+\frac{x}{n})}{n+1}\right]^{n+1}\\
& = \left[1+\frac{x}{n(n+1)}\right]^{n+1}\\
& \leq \left[1+\frac{x}{n+1}\right]^{n+1}
\end{aligned}
\end{equation}
The first inequality holds for the arithmetic mean equality. We could deduce that $(1+\frac{x}{n})^{n} \leq e^{x}$. Therefore, we have:
\begin{equation}
J_{\pi_{t}}(\theta) \log \frac{J_{\pi_{*}}(\theta)}{J_{\pi_{t}}(\theta)} \leq \epsilon
\end{equation}
Based on Assumption \ref{assumption4}, we could deduce the upper bound of average discounted reward. Then the lemma could be proved.
\end{proof}
\end{lemma}

After bounding the log ratio between the expected optimal average reward and the instantaneous reward. We could then move to the bound of the instantaneous posterior weight $W_{t_{k}}$. Which is a crucial factor in the following proving process. At each time step, the posterior weight will be updated based on the previous policy and the observed experiment process. First we define the counter function $N_{\pi}(t):=\sum_{t=0}^{t-1} \sum_{\pi \in \Pi} \mathbbm{1}\left\{\pi_{e(t)}=\pi\right\}$ as the total number of the time instants during the period of $t$ when policy $\pi$ was conducted. When Assumption \ref{assumption3} holds, we could bound the posterior weight based on the count function in $k$ epoch and the average transition time $\tilde \tau$.

\begin{lemma}
\label{lemma3}
Under Assumption \ref{assumption3}, for each stationary near-optimal policy $\pi$ and epoch counter $k \geq 1$. The following upper bound holds for negative log-density.
$$
-\log W_{t_{k}}(\pi) \leq \frac{\epsilon}{\Gamma}|S|^{2}(  \rho(k_{\pi}) \sqrt{k_{\pi}} + k_{\pi} \tilde{\tau}_{t_{k},k_{\pi}})
$$
\begin{proof}
When $W_{t_{k}} \leq 1$, we could have:
\begin{equation}
W_{t_{k}}(\theta):=\exp \sum_{\pi, s_{1}, s_{2}} \mathcal{H}\left(N_{\pi}(k), \pi\right) \log \frac{J_{\pi_{t}}(\theta)}{J_{\pi_{*}}(\theta)}
\end{equation}
Based on the definition of the counter $\mathcal{H}$, we could deduce the value of the posterior weight in a single epoch:
\begin{equation}
\begin{aligned}
&W_{t_{k}}(\theta)\\
&= \exp \left(\sum_{t=1}^{\infty} \mathbbm{1}\left\{\pi_{e(t)}=\pi,\left(S_{t}, S_{t+1}\right)=\left(s_{1}, s_{2}\right)\right\}\log \frac{J_{\pi_{t}}(\theta)}{J_{\pi_{*}}(\theta)}\right)\\
& = \exp \left(\sum_{\pi \in \Pi} \sum_{\left(s_{1}, s_{2}\right) \in \mathcal{S}^{2}} \sum_{t=1}^{T} \mathbbm{1}\left\{\pi_{e(i)}=\pi,\left(S_{t}, S_{t+1}\right)=\left(s_{1}, s_{2}\right)\right\}\right.\\
&\left.\quad \log \frac{J_{\pi_{t}}(\theta)}{J_{\pi_{*}}(\theta)}\right)\\
& = \exp \left(N_{\pi}(t) \sum_{\left(s_{1}, s_{2}\right) \in \mathcal{S}^{2}} \sum_{t=0}^{t-1} \frac{\mathbbm{1}\left\{\pi_{e(t)}=\pi,\left(S_{t}, S_{t+1}\right)=\left(s_{1}, s_{2}\right)\right\}}{N_{\pi}(t)}\right.\\
&\left.\quad \log \frac{J_{\pi_{t}}(\theta)}{J_{\pi_{*}}(\theta)}\right)\\
\end{aligned}
\end{equation}
Where $N_{\pi}(t):=\sum_{t=0}^{t-1}\sum_{\pi \in \Pi} \mathbbm{1}\left\{\pi_{e(t)}=\pi\right\}$ represents the total number of the time instants during the period of $t$ when policy $\pi$ was conducted.

When Assumption \ref{assumption3} holds, we could know that $N_{\pi}(t) = \tilde{\tau}_{\pi_{t_{k}},N_{\pi}(k)}$, where $N_{\pi}(k) :=\sum_{k=0}^{K}\sum_{\pi \in \Pi} \mathbbm{1}\left\{\pi_{e(k)}=\pi\right\} $ holds for the number of the epochs where policy $\pi$ was chosen(The notation of $\tau$ will be represented as $N_{\pi}(k) = k_{\pi}$, $\tilde{\tau}_{\pi_{t_{k}},N_{\pi}(k)} = \tilde{\tau}_{t_{k},k_{\pi}}$). Therefore, we could have:
\begin{equation}
\begin{aligned}
& -\log W_{t_{k}}(\pi)\\
& = -N_{\pi}(t) \sum_{\left(s_{1}, s_{2}\right) \in \mathcal{S}^{2}} \sum_{t=0}^{t-1} \frac{\mathbbm{1}\left\{\pi_{e(t)}=\pi,\left(S_{t}, S_{t+1}\right)=\left(s_{1}, s_{2}\right)\right\}}{N_{\pi}(t)}\log \frac{J_{\pi_{t}}(\theta)}{J_{\pi_{*}}(\theta)}\\
& = -\sum_{\left(s_{1}, s_{2}\right) \in \mathcal{S}^{2}} \tilde{\tau}_{t_{k},k_{\pi}}  \mathcal{H}_{\left(s_{1}, s_{2}\right)}\left(\tilde{\tau}_{t_{k},k_{\pi}}, \pi\right) \log \frac{J_{\pi_{t}}(\theta)}{J_{\pi_{*}}(\theta)}\\
& = \sum_{\left(s_{1}, s_{2}\right) \in \mathcal{S}^{2}} \left[\tilde{\tau}_{t_{k},k_{\pi}}  \mathcal{H}_{\left(s_{1}, s_{2}\right)}\left(\tilde{\tau}_{t_{k},k_{\pi}}, \pi\right) - k_{\pi}\tilde{\tau}_{t_{k},k_{\pi}}\theta_{\pi}(s_{1}|s_{2})\right]\log \frac{J_{\pi_{*}}(\theta)}{J_{\pi_{t}}(\theta)}\\
& + \sum_{\left(s_{1}, s_{2}\right) \in \mathcal{S}^{2}} k_{\pi}\tilde{\tau}_{t_{k},k_{\pi}}\theta(s_{1}|s_{2}) \log \frac{J_{\pi_{*}}(\theta)}{J_{\pi_{t}}(\theta)}\\
\end{aligned}
\end{equation}
The last equation is based on the logarithmic property $\log \frac{A}{B} = - \log \frac{B}{A}$. Based on the Assumption \ref{assumption3}, define $\rho(x) := O(\sqrt{\log \log(x)})$.
\begin{equation}
\begin{aligned}
- \log W_{t_{k}}(\pi)& \leq \sum_{\left(s_{1}, s_{2}\right) \in \mathcal{S}^{2}} \rho(k_{\pi}) \sqrt{k_{\pi}} \log \frac{J_{\pi_{*}}(\theta)}{J_{\pi_{t}}(\theta)}\\
&+ k_{\pi}\tilde{\tau}_{t_{k},k_{\pi}} \sum_{\left(s_{1}, s_{2}\right) \in \mathcal{S}^{2}}\theta(s_{1}|s_{2}) \log \frac{J_{\pi_{*}}(\theta)}{J_{\pi_{t}}(\theta)}\\
& \leq \frac{\epsilon}{\Gamma}|S|^{2}(  \rho(k_{\pi}) \sqrt{k_{\pi}} + k_{\pi} \tilde{\tau}_{t_{k},k_{\pi}})
\end{aligned}
\end{equation}
\end{proof}
\end{lemma}

A proper optimization method should lead to promised margin between the expected discounted reward and the real reward. In order to achieve such results, numerous amount of iteration will be conducted. Therefore, from the astringency proof we proposed in section \ref{section4.1}, we could deduce the bound of the expected convergence time during the optimization process. In Lemma \ref{lemma3.3}, we give the bound the instantaneous difference between the real reward and the expected reward with $\sqrt{T}$. This bound is inversely proportional to $\sqrt{T}$ since our update method updates towards optimal direction(Lemma \ref{lemma0}). For the sake of brevity, the full proof will be shown in Appendix \ref{lemma3.3replacement}.
\begin{lemma}
\label{lemma3.3}
The difference between the local optimal average reward and the instantaneous average reward could be bounded by:
$$
|J_{\pi_{t}}-J^{*}| \leq \tilde{\mathcal{O}}(\frac{1}{\sqrt{T}})
$$
\begin{proof}
We could know that the current policy probability distribution is updated based on the previous distribution and the current optimal policy distribution:
\begin{equation}
\mu_{t}(\pi) = W_{t}\mu_{t}(\pi) + (1-W_{t})\pi_{t}^{*}(s,\theta_{t_{k}}
\end{equation}
We could extend this result to reward function:
\begin{equation}
\label{equation500}
\begin{aligned}
J_{\pi_{t}}&=W_{t} J_{\pi_{t-1}}+\left(1-W_{t}\right) J^{*}\left(\theta_{t}\right)\\
J_{\pi_{t}}^{2} &=W_{t}^{2} J_{\pi{t-1}}^{2}+\left(1-W_{t}\right)^{2} J^{*^{2}}+2 W_{t}\left(1-W_{t}\right) J^{*} J_{\pi_{t-1}} \\
& \leq W_{t}^{2} J_{\pi t}^{2}+\left(1-W_{t}\right)^{2} J^{* 2}+2 W_{t}\left(1-W_{t}\right) J^{*} J_{\pi_{t}}
\end{aligned}
\end{equation}
The inequality is based on the monotonicity of the algorithm. We could simplify Equation \ref{equation500}:
\begin{equation}
\begin{aligned}
\left(1-W_{t}^{2}\right) J_{\pi_{t}}^{2} &\leq \left(1-W_{t}\right)^{2} J^{*^{2}}+2 W_{t}\left(1-W_{t}\right) J^{*} J_{\pi_{t}} \\
\left(1+W_{t}\right) J_{\pi t}^{2} &\leq \left(1-W_{t}\right) J^{*^{2}}+2 W_{t} J^{*} J_{\pi_{t}} \\
J_{\pi_{t}}^{2}+W_{t} J_{\pi_{t}}^{2} &\leq J^{*^{2}}-W_{t} J^{*^{2}}+2 W_{t} J^{*} J_{\pi_{t}}\\
W_{t}\left(J_{\pi_{t}}^{2}+J^{*^{2}}\right) &\leq J^{*^{2}}-J_{\pi_{t}}^{2}+2 W_{t} J^{*} J_{\pi_{t}}\\
J_{\pi_{t}}^{2}+J^{*^{2}} &\leq \frac{1}{W_{t}}\left(J^{*^{2}}-J_{\pi_{t}}^{2}\right)+2 J^{*} J_{\pi_{t}}
\end{aligned}
\end{equation}
Based on the definition of the regret of each time step, we could deduce the bound of the instantaneous regret:
\begin{equation}
\begin{aligned}
\left(J_{\pi t}-J^{*}\right)^{2} &=J_{\pi_{t}}^{2}+J^{*^{2}}-2 J_{\pi t} J^{*} \\
& \leq \frac{1}{w_{t}}\left(J^{*^{2}}-J_{\pi_{t}}^{2}\right)+2 J^{*} J_{\pi_{t}}-2 J_{\pi_{t}} J^{*} \\
&=\frac{1}{w_{t}}\left(J^{*^{2}}-J_{\pi_{t}}^{2}\right) \\
&=\frac{1}{w_{t}}\left(J^{*}-J_{\pi_{t}}\right)\left(J^{*}+J_{\pi_{t}}\right)
\end{aligned}
\end{equation}

\begin{equation}
|J_{\pi_{t}}-J^{*}| \leq \frac{1}{W_{t}}|J_{\pi_{t}}+J^{*}|
\end{equation}
From Lemma \ref{lemma3}, we could know that $-\log W_{t_{k}}(\pi)$ is bounded by $B$, with $B = \frac{\epsilon}{\Gamma}|S|^{2}(\rho(k_{\pi}) \sqrt{k_{\pi}} + k_{\pi} \tilde{\tau}_{t_{k},k_{\pi}})$. Therefore, we could construct the following inequalities.
\begin{equation}
\begin{aligned}
W_{t_{k}}-1 &\geq \log W_{t_{k}} \geq -B\\
\frac{1}{W_{t_{k}}} &\leq \frac{1}{1-B}
\end{aligned}
\end{equation}
Factor $B$ is proportional to parameter $k_{\pi}$ which could be bounded by the total number of episode of under total time $T$. Therefore, we could bound $\frac{1}{W_{t}}$ by $T$(Ignoring the constants):
\begin{equation}
\begin{aligned}
\frac{1}{W_{t}} &\leq \frac{1}{1-\frac{\epsilon}{\Gamma}|S|^{2}(\rho(k_{\pi}) \sqrt{k_{\pi}} + k_{\pi} \tilde{\tau}_{t_{k},k_{\pi}})}\\
&\leq \frac{1}{1-\sqrt{\sqrt{T}}-\sqrt{T}}
\end{aligned}
\end{equation}
Based on Assumption \ref{assumption4}, the average discounted reward function is bounded by $\Gamma$. So the difference between the local optimal average reward and the instantaneous average reward could be bounded by:
\begin{equation}
\begin{aligned}
|J_{\pi_{t}}-J^{*}| &\leq \frac{1}{W_{t}}|J^{*}+J_{\pi_{t}|}\\
&\leq \frac{2}{1-\frac{\epsilon}{\Gamma}|S|^{2\Gamma}(\rho(k_{\pi}) \sqrt{k_{\pi}} + k_{\pi} \tilde{\tau}_{t_{k},k_{\pi}})}\\
&\leq \tilde{\mathcal{O}} (\frac{2\Gamma^{2}}{S^{2}\sqrt{T}})
\end{aligned}
\end{equation}
\end{proof}
\end{lemma}

Therefore, we could combine the previous Lemmas together to get the final regret bound of $Reg_{T}^{1}$.

\begin{theorem}
\label{theorem4}
The first part of the regret in time step $T$ is bounded by:
$$
Reg_{T}^{1} \leq \tilde{\mathcal{O}}(\frac{\sqrt{T}}{S^{2}})
$$
\begin{proof}
From the definition before, we could know that $Reg_{T}^{1}$ could be represented as:
\begin{equation}
Reg_{T}^{1} = TJ_{\pi_{\hat{k}}}(\hat \theta) - \sum_{t = 1}^{T}J_{\pi_{t}}(\hat \theta)
\end{equation}
Since this theorem won't involve the transformation of the transition probability. So let $J_{\pi}(\theta) = J_{\pi}$. Based on the update rule of the posterior distribution $\mu_{t+1}(\pi)$ of policy $\pi$. We could divide the average discounted return into several parts:\\
At time step $t = T$, we could assume the instantaneous regret equals to zero:
\begin{equation}
Reg_{t_{T}}^{1} = J_{\pi_{\hat k}}- J_{\pi_{T}} = 0
\end{equation}
At time step $t = T-1$, define the local optimal average discounted return as $J_{\pi}^{*}$. Note that this local optimal value is virtual. The instantaneous regret could be represented as:
\begin{equation}
\begin{aligned}
Reg_{t_{T-1}}^{1} &= J_{\pi_{\hat k}} - J_{\pi_{T-1}}\\
&= W_{t-1}J_{\pi_{T-1}} + (1-W_{t-1})J_{\pi}^{*} - J_{\pi_{T-1}}\\
&= (W_{t-1}-1)J_{\pi_{T-1}} + (1-W_{t-1})J_{\pi}^{*}\\
&= (1-W_{t-1})(J_{\pi}^{*} - J_{\pi_{T-1}})
\end{aligned}
\end{equation}
In a similar fashion, at time step $t = T-2$, the instantaneous regret could be represented as:
\begin{equation}
\begin{aligned}
Reg_{t_{T-2}}^{1} &= J_{\pi_{\hat k}} - J_{\pi_{T-2}}\\
&= J_{\pi_{\hat k}} - J_{\pi_{T-1}} + J_{\pi_{T-1}} - J_{\pi_{T-2}}\\
&= (1-W_{t-1})(J_{\pi}^{*} - J_{\pi_{T-1}}) + (1-W_{t-2})(J_{\pi}^{*} - J_{\pi_{T-1}})
\end{aligned}
\end{equation}
Based on Lemma \ref{lemma3.3}, the difference between the local optimal value and the current average return could be bounded by:
\begin{equation}
|J_{\pi}^{*} - J_{\pi_{t}}| \leq \tilde{\mathcal{O}}(\frac{1}{\sqrt{T}})
\end{equation}

The sub-optimal models are sampled when their posterior probability is larger than $\frac{1}{T}$. This ensures the time complexity of the Thompson sampling process is no more than $O(1)$. So we could deduce the total regret in time step $T$.
\begin{equation}
\begin{aligned}
Reg_{T}^{1} &= \frac{1}{T} (Reg_{t_{T-1}}^{1} + Reg_{t_{T-2}}^{1} + \cdots + Reg_{t_{1}}^{1})\\
&\leq \tilde{\mathcal{O}}(\frac{2\Gamma^{2}}{S^{2}\sqrt{T}})(\frac{T-1}{T} + \frac{T-2}{T} + \cdots \frac{1}{T})\\
&\leq \tilde{\mathcal{O}}(\frac{\Gamma^{2}\sqrt{T}}{S^{2}})
\end{aligned}
\end{equation}
\end{proof}
\end{theorem}

After giving the first part of the total regret bound, we then move to the proof of the second and third part. Based on the definition in Equation \ref{equation17}, the second regret bound is mainly related to the difference between transition probability. So we could denote the average discounted return as $J_{\pi_{t}}(\hat \theta) = J(\hat \theta)$, $J_{\pi_{t}}(\theta_{t}) = J(\theta_{t})$. Note that the $\sum_{t=1}^{T}J(\hat \theta)$ is a virtual value which represents the average discounted return in time step $T$ with transition probability $\hat \theta$. From the previous definition of the Bellman iterator of the average discounted return, we could deduct the bound of $Reg_{T}^{2}$(The full proof will be shown in the appendix for the sake of brevity):
\begin{theorem}
\label{theorem5}
The second part of the regret in time step $T$ is bounded by:
$$
Reg_{T}^{2} \leq \tilde{\mathcal{O}}( D\sqrt{SAT} )
$$

%Since the transition probability $\hat \theta$ and $\theta_{t}$ are identically distributed on the history. That means the  for any measurable function $f$ and any $h_{t_{k}}$-measurable random variable $X$, we have $\mathbb{E}\left[f\left(\hat \theta, X\right) \mid h_{t_{k}}\right]=\mathbb{E}\left[f\left(\theta_{k}, X\right) \mid h_{t_{k}}\right]$. Therefore, we could deduce the relationship between average discounted return under optimal transition probability and near-optimal transition probability.

\end{theorem}
Finally, for the last part of the regret bound. $Reg_{T}^{3}$ is calculated by the difference between the instantaneous virtual average reward and the real reward. In Theorem \ref {theorem6}, we decompose the regret into two parts $Y_{t}^{1}$ and $Y_{t}^{2}$. We then use the Azuma-Hoeffding's inequality to bound $Y_{t}^{1}$ and $Y_{t}^{2}$ respectively. Therefore, we could get the regret bound of $Reg_{T}^{3}$(The full proof will be shown in the appendix):
\begin{theorem}
\label{theorem6}
The third part of the regret in time step $T$ is bounded by:
$$
Reg_{T}^{3} \leq \tilde{\mathcal{O}}(D\sqrt{ST})
$$
\end{theorem}

\section{Conclusion}
In this paper, we propose a policy-based posterior optimization method that achieves the best total regret bound $\tilde{\mathcal{O}}(\Gamma^{2}\sqrt{T}/S^{2})$ in finite-horizon stochastic game. This algorithm provides a new vision on the trade-off problem between exploration and exploitation by solving a posterior update problem. The posterior update problem could be solved by balancing between long-term policy and current greedy policy. Our research results shows that this posterior sampling method outperforms other optimization algorithms both theoretically and empirically.

In the future work, we aim to extend the application scope of our algorithm to continuous space. Sampling method had been proved to be efficient in discrete environment. But it still occurs many obstacles in this area. Our approach solves the discrete problems with count-based posterior weight. Such idea could be transplanted to continuous environment as well. We could represent the difference between the state of the continuous spaces with specific metric. Then adopt our method in such environment.
\bibliography{example_paper}

\begin{thebibliography}{16}
\providecommand{\natexlab}[1]{#1}
\providecommand{\url}[1]{\texttt{#1}}
\expandafter\ifx\csname urlstyle\endcsname\relax
  \providecommand{\doi}[1]{doi: #1}\else
  \providecommand{\doi}{doi: \begingroup \urlstyle{rm}\Url}\fi

\bibitem[Auer(2002)]{auer2002using}
Auer, P.
\newblock Using confidence bounds for exploitation-exploration trade-offs.
\newblock \emph{Journal of Machine Learning Research}, 3\penalty0
  (Nov):\penalty0 397--422, 2002.

\bibitem[Brafman \& Tennenholtz(2003)Brafman and
  Tennenholtz]{10.1162/153244303765208377}
Brafman, R.~I. and Tennenholtz, M.
\newblock R-max - a general polynomial time algorithm for near-optimal
  reinforcement learning.
\newblock \emph{J. Mach. Learn. Res.}, 3\penalty0 (null):\penalty0 213–231,
  mar 2003.
\newblock ISSN 1532-4435.
\newblock \doi{10.1162/153244303765208377}.

\bibitem[Chapelle \& Li(2011)Chapelle and Li]{chapelle2011empirical}
Chapelle, O. and Li, L.
\newblock An empirical evaluation of thompson sampling.
\newblock \emph{Advances in neural information processing systems},
  24:\penalty0 2249--2257, 2011.

\bibitem[Cui \& Yang(2021)Cui and Yang]{Cui2021MinimaxSC}
Cui, Q. and Yang, L.~F.
\newblock Minimax sample complexity for turn-based stochastic game.
\newblock In \emph{UAI}, 2021.

\bibitem[Hao et~al.(2019)Hao, Abbasi-Yadkori, Wen, and
  Cheng]{Hao2019BootstrappingUC}
Hao, B., Abbasi-Yadkori, Y., Wen, Z., and Cheng, G.
\newblock Bootstrapping upper confidence bound.
\newblock In \emph{NeurIPS}, 2019.

\bibitem[Kveton et~al.(2020)Kveton, Zaheer, Szepesvari, Li, Ghavamzadeh, and
  Boutilier]{Kveton2020RandomizedEI}
Kveton, B., Zaheer, M., Szepesvari, C., Li, L., Ghavamzadeh, M., and Boutilier,
  C.
\newblock Randomized exploration in generalized linear bandits.
\newblock In \emph{AISTATS}, 2020.

\bibitem[Lai \& Robbins(1985)Lai and Robbins]{lai1985asymptotically}
Lai, T.~L. and Robbins, H.
\newblock Asymptotically efficient adaptive allocation rules.
\newblock \emph{Advances in applied mathematics}, 6\penalty0 (1):\penalty0
  4--22, 1985.

\bibitem[Osband et~al.(2013)Osband, Russo, and Van~Roy]{osband2013more}
Osband, I., Russo, D., and Van~Roy, B.
\newblock (more) efficient reinforcement learning via posterior sampling.
\newblock \emph{arXiv preprint arXiv:1306.0940}, 2013.

\bibitem[Ouyang et~al.(2017)Ouyang, Gagrani, Nayyar, and
  Jain]{10.5555/3294771.3294898}
Ouyang, Y., Gagrani, M., Nayyar, A., and Jain, R.
\newblock Learning unknown markov decision processes: A thompson sampling
  approach.
\newblock In \emph{Proceedings of the 31st International Conference on Neural
  Information Processing Systems}, NIPS'17, pp.\  1333–1342, Red Hook, NY,
  USA, 2017. Curran Associates Inc.
\newblock ISBN 9781510860964.

\bibitem[Russo \& Van~Roy(2014)Russo and Van~Roy]{russo2014learning}
Russo, D. and Van~Roy, B.
\newblock Learning to optimize via posterior sampling.
\newblock \emph{Mathematics of Operations Research}, 39\penalty0 (4):\penalty0
  1221--1243, 2014.

\bibitem[Szita \& Lörincz(2009)Szita and Lörincz]{inproceedings}
Szita, I. and Lörincz, A.
\newblock Optimistic initialization and greediness lead to polynomial time
  learning in factored mdps.
\newblock volume 382, pp.\  126, 06 2009.
\newblock \doi{10.1145/1553374.1553502}.

\bibitem[Thompson(1933)]{thompson1933likelihood}
Thompson, W.~R.
\newblock On the likelihood that one unknown probability exceeds another in
  view of the evidence of two samples.
\newblock \emph{Biometrika}, 25\penalty0 (3/4):\penalty0 285--294, 1933.

\bibitem[Tokic(2010)]{tokic2010adaptive}
Tokic, M.
\newblock Adaptive $\varepsilon$-greedy exploration in reinforcement learning
  based on value differences.
\newblock In \emph{Annual Conference on Artificial Intelligence}, pp.\
  203--210. Springer, 2010.

\bibitem[{Wal, van der}(1977)]{72bf2cf9895a47e7be9a668a25215110}
{Wal, van der}, J.
\newblock \emph{Successive approximation for average reward Markov games}.
\newblock Memorandum COSOR. Technische Hogeschool Eindhoven, 1977.

\bibitem[Wei et~al.(2017)Wei, Hong, and Lu]{NIPS2017_36e729ec}
Wei, C.-Y., Hong, Y.-T., and Lu, C.-J.
\newblock Online reinforcement learning in stochastic games.
\newblock In Guyon, I., Luxburg, U.~V., Bengio, S., Wallach, H., Fergus, R.,
  Vishwanathan, S., and Garnett, R. (eds.), \emph{Advances in Neural
  Information Processing Systems}, volume~30. Curran Associates, Inc., 2017.
\newblock URL
  \url{https://proceedings.neurips.cc/paper/2017/file/36e729ec173b94133d8fa552e4029f8b-Paper.pdf}.

\bibitem[Wei et~al.(2020)Wei, Jahromi, Luo, Sharma, and Jain]{wei2020model}
Wei, C.-Y., Jahromi, M.~J., Luo, H., Sharma, H., and Jain, R.
\newblock Model-free reinforcement learning in infinite-horizon average-reward
  markov decision processes.
\newblock In \emph{International conference on machine learning}, pp.\
  10170--10180. PMLR, 2020.

\end{thebibliography}
\bibliographystyle{icml2022}
\nocite{tokic2010adaptive}
\nocite{inproceedings}
\nocite{lai1985asymptotically}
\nocite{auer2002using}
\nocite{Hao2019BootstrappingUC}
\nocite{thompson1933likelihood}
\nocite{osband2013more}
\nocite{chapelle2011empirical}
\nocite{russo2014learning}
\nocite{NIPS2017_36e729ec}
\nocite{wei2020model}
\nocite{10.5555/3294771.3294898}
\nocite{Cui2021MinimaxSC}
\nocite{Kveton2020RandomizedEI}
\nocite{72bf2cf9895a47e7be9a668a25215110}

%%%%%%%%%%%%%%%%%%%%%%%%%%%%%%%%%%%%%%%%%%%%%%%%%%%%%%%%%%%%%%%%%%%%%%%%%%%%%%%
%%%%%%%%%%%%%%%%%%%%%%%%%%%%%%%%%%%%%%%%%%%%%%%%%%%%%%%%%%%%%%%%%%%%%%%%%%%%%%%
% APPENDIX
%%%%%%%%%%%%%%%%%%%%%%%%%%%%%%%%%%%%%%%%%%%%%%%%%%%%%%%%%%%%%%%%%%%%%%%%%%%%%%%
%%%%%%%%%%%%%%%%%%%%%%%%%%%%%%%%%%%%%%%%%%%%%%%%%%%%%%%%%%%%%%%%%%%%%%%%%%%%%%%
\newpage
\appendix
\onecolumn
\section{The Convergence of PI}

\begin{lemma}
Under update algorithm PPI, the average discounted return should be monotonically increased.
\begin{proof}
From Algorithm \ref{alg2}, we could deduce the update rule of the average discounted return:
\begin{equation}
\begin{aligned}
J_{\pi_{t}}(\theta) - J_{\pi_{t-1}}(\theta) &= (W_{t}-1)J_{t-1}(\theta) + (1-W_{t})J_{\pi^{*}}(s,\theta)\\
&= (1-W_{t})(J_{\pi^{*}}(s,\theta)-J_{\pi_{t-1}}(\theta))
\end{aligned}
\end{equation}
When $J_{\pi^{*}}(s,\theta) \geq J_{\pi_{t-1}}(\theta)$, we could deduce that $\log\frac{J_{\pi_{t}}(\theta)}{J_{\pi^{*}}(s,\theta)} \leq 1$. So the posterior weight $W_{t}$ is less than 1. This result holds vice versa. The first term $1-W_{t} \leq 0$ when $J_{\pi^{*}}(s,\theta) \leq J_{\pi_{t-1}}(\theta)$. Therefore, we could prove that:
\begin{equation}
J_{\pi_{t}}(\theta) - J_{\pi_{t-1}}(\theta) = (1-W_{t})(J_{\pi^{*}}(s,\theta)-J_{\pi_{t-1}}(\theta)) \geq 0
\end{equation}
The sequence $J_{\pi_{t}}(\theta)$ is monotonically increased with time step $t$.
\end{proof}
\end{lemma}

\begin{lemma}
\label{lemma2}
Suppose Assumption \ref{assumption1} and Assumption \ref{assumption2} hold for some stochastic games $M$. Let $v_{i}$ be the state value in iteration $i$. Define $N$ as the maximum iteration number of the algorithm. Then $\pi_{t_{k}}$ is $\epsilon$-optimal after $N$ iterations.
\begin{proof}
Define $D = \min_{s}\{\mu_{i+1}(\pi)-\mu_{\pi}\}$ and $U = \max_{s}\{\mu_{i+1}(\pi)-\mu_{i}(\pi)\}$. Then we could deduce:
\begin{equation}
\begin{aligned}
D + \mu_{N}(\pi) &\leq \mu_{N+1}\\
&\leq W_{i}\mu_{N} + (1-W_{i})\pi_{i}^{*}(s,\theta)\\
&\leq W_{i}\mu_{N} + (1-W_{i})(r_{N} + \theta_{} v_{N})
\end{aligned}
\end{equation}
Since $0 \textless W_{i} \textless 1$, the upper equation could be turned to:
\begin{equation}
D \leq (1-W_{i})J_{\pi_{i}}(\theta)
\end{equation}
Let $\pi^{*}$ be the optimal policy under all states that satisfies $\pi^{*} := \sum_{s \in S}\pi_{i}^{*}(s,\theta)$. Then
\begin{equation}
D \leq (1-W_{i})J_{\pi_{i}}(\theta) \leq (1-W_{i})J_{\pi^{*}}(\theta)
\end{equation}
In a similar way, we could also prove $U \geq (1-W_{i})J_{\pi^{*}}(\theta)$. From the definition of the stopping criterion of the Policy Iteration algorithm, we could assume $U - D \leq (1-W_{i})\epsilon$. Therefore, we have
\begin{equation}
\begin{aligned}
U &\leq D + (1-W_{i})\gamma \\
&\leq (1-W_{i})J_{\pi_{i}}(\theta)+(1-W_{i})\epsilon\\
&\leq (1-W_{i})(J_{\pi_{i}}(\theta)+\epsilon)\\
(1-W_{i})J_{\pi^{*}} &\leq (1-W_{i})(J_{\pi_{i}}(\theta)+\epsilon)\\
J_{\pi^{*}} &\leq J_{\pi_{i}}(\theta)+\epsilon
\end{aligned}
\end{equation}
We could deduce that stationary policy $\pi$ is $\epsilon$-optimal after $N$ iterations.
\end{proof}
\end{lemma}

\section{Regret Bound Analysis}
To analyze our algorithm's performance over $T$ time step. We define the number of macro episodes $M = \mathbbm{1}\{t_{k} \leq T\}$. An episode is defined as the set of the time steps under stopping criterions. Therefore, we could deduce the bound of the number of episode.
\begin{lemma}
\label{lemma1.1}
Under the stopping criterion, the number of episodes $M$ could be bounded by:
$$
M \leq S A \log (T)
$$
\begin{proof}
The stopping criterion is triggered whenever the visits number of the initial state-action pair is doubled. So $M$ could be represented as:
\begin{equation}
M_{(s, a)}=\left\{k \leq K_{T}: N_{t_{k}}(s, a)>2 N_{t_{k-1}}(s, a)\right\}
\end{equation}
Since the number of the visit to state-action pair $(s,a)$ is doubled at the beginning of every epoch $k$. The size of $\mathcal{M}_{(s, a)}$ should be no larger than $O(\log (T))$. Assume $\left|\mathcal{M}_{(s, a)}\right| \geq \log \left(N_{T+1}(s, a)\right)+1$. We could have:
\begin{equation}
\begin{aligned}
N_{t_{K_{T}}}(s, a) &= \prod_{k \leq K_{T}, N_{t_{k-1}}(s, a) \geq 1} \frac{N_{t_{k}}(s, a)}{N_{t_{k-1}}(s, a)}\\
&> \prod_{k \in \mathcal{M}_{(s, a)}, N_{t_{k-1}}(s, a) \geq 1} 2 \\
&\geq N_{T+1}(s, a)
\end{aligned}
\end{equation}
This contradicts the fact that $N_{t_{K_{T}}}(s, a) \leq N_{T+1}(s, a)$. This leads to $\left|\mathcal{M}_{(s, a)}\right| \leq \log \left(N_{T+1}(s, a)\right)$. Therefore, we could obtain the bound of the number of the episodes:
\begin{equation}
\begin{aligned}
M & \leq 1+\sum_{(s, a)}\left|\mathcal{M}_{(s, a)}\right|\\
&\leq 1+\sum_{(s, a)} \log \left(N_{T+1}(s, a)\right) \\
& \leq 1+S A \log \left(\sum_{(s, a)} N_{T+1}(s, a) / S A\right)\\
&=1+S A \log (T / S A)
\end{aligned}
\end{equation}
Since the logarithmic function is concave, we could simplify the inequality to:
\begin{equation}
M \leq S A \log(T)
\end{equation}
\end{proof}
\end{lemma}

\begin{lemma}
\label{lemma1.2}
The total number of episodes of total time step $T$ could be bounded by:
$$
K_{T} \leq \sqrt{2 S A T \log (T)}
$$
\begin{proof}
Define macro episodes with start times $t_{n_{i}},i=1,2,\cdots$ where $t_{n_{1}}=t_{1}$,we could have
$$
t_{n_{i+1}}=\min \left\{t_{k}>t_{n_{i}}: N_{t_{k}}(s, a)>2 N_{t_{k-1}}(s, a) \right\}
$$
Let $\tilde{T}_{i}=\sum_{k=n_{i}}^{n_{i+1}-1} T_{k}$ be the length of the ith episode. Therefore, within the $i$th macro episode, $T_{k} = T_{k-1}+1$ for all $k = n_{i},n_{i}+1,\cdots,n_{i+1}-2$.
\begin{equation}
\begin{aligned}
\tilde{T}_{i}&=\sum_{k=n_{i}}^{n_{i+1}-1} T_{k} \\ &=\sum_{j=1}^{n_{i+1}-n_{i}-1}\left(T_{n_{i}-1}+j\right)+T_{n_{i+1}-1} \\
& \geq \sum_{j=1}^{n_{i+1}-n_{i}-1}(j+1)+1=0.5\left(n_{i+1}-n_{i}\right)\left(n_{i+1}-n_{i}+1\right) .
\end{aligned}
\end{equation}
Consequently,$n_{i+1}-n_{i} \leq \sqrt{2 \tilde{T}_{i}}$, for all $i = 1,\cdots,M$. From this property, we could obtain:
\begin{equation}
\label{equation29}
K_{T}=n_{M+1}-1=\sum_{i=1}^{M}\left(n_{i+1}-n_{i}\right) \leq \sum_{i=1}^{M} \sqrt{2 \tilde{T}_{i}}
\end{equation}
Based on Equation \ref{equation29} and $\sum_{i=1}^{M} \tilde{T}_{i}=T$, we could get:
\begin{equation}
K_{T} \leq \sum_{i=1}^{M} \sqrt{2 \tilde{T}_{i}} \leq \sqrt{M \sum_{i=1}^{M} 2 \tilde{T}_{i}}=\sqrt{2 M T}
\end{equation}
Where the second inequality is based on Cauchy-Schwarz inequality. From Lemma \ref{lemma1.1}, we could know that the number of the macro episodes until time $T$ is bounded by $M \leq SA\log(T)$. Therefore, the lemma could be proved.
\end{proof}
\end{lemma}

\begin{theorem}
The regret that generated from each epoch $k$ could be bounded by
$$
Reg_{K} \leq \tilde{\mathcal{O}}(\sqrt{2\sqrt{SAT}})
$$
with high probability.
\begin{proof}
In order to bound the total regret in epoch $k$, we begin by proving the average regret bound.
\begin{equation}
\begin{aligned}
||J_{\pi_{*}}(\theta_{*}) - \frac{1}{K}\sum_{k=1}^{K}r_{\pi_{\hat k}} (s_{k},a_{k})||&\leq ||J_{\pi_{*}}(\theta_{*}) - J_{\pi_{\hat k}}(\hat \theta_{k})||\\
&\leq ||J_{\pi_{*}}(\theta_{*}) - J_{\pi_{*}}(\hat \theta_{k})|| + ||J_{\pi_{*}}(\hat \theta_{k}) - J_{\pi_{\hat k}}(\hat \theta_{k})||\\
\end{aligned}
\end{equation}
Based on the previous definition for average discounted return $J$(Equation \ref{equation1}). We could deduce its Bellman operator $J_{\pi_{t}}(\theta(t)) = r(s,a)+\gamma J_{\pi(t-1)}(\theta(t-1))$.Then the upper inequality could be altered to:
\begin{equation}
\begin{aligned}
||J_{\pi_{*}}(\theta_{*}) - J_{\pi_{\hat k}}(\hat \theta_{k})||&\leq ||J_{\pi_{*}}(\theta_{*}) - J_{\pi_{*}}(\hat \theta_{k})|| + ||J_{\pi_{*}}(\hat \theta_{k}) - J_{\pi_{\hat k}}(\hat \theta_{k})||\\
&\leq \gamma||\theta_{*}J_{\pi_{*}}(\theta_{*}) - \hat \theta_{k}J_{\pi_{*}}(\theta_{*})|| + \gamma||J_{\pi_{*}}(\theta_{*}) - J_{\pi_{\hat{k}}}(\hat \theta_{k})||
\end{aligned}
\end{equation}
Define $\beta = \frac{1}{1-\gamma}$. We then subtract the second term $\gamma||J_{\pi_{*}}(\theta_{*}) - J_{\pi_{\hat{k}}}(\hat \theta_{k})||$ to the left side of the inequality:
\begin{equation}
\label{equation16}
\begin{aligned}
(1-\gamma)||J_{\pi_{*}}(\theta_{*}) - J_{\pi_{\hat k}}(\hat \theta_{k})|| &\leq \gamma||\theta_{*}J_{\pi_{*}}(\theta_{*}) - \hat \theta_{k}J_{\pi_{*}}(\theta_{*})||\\
||J_{\pi_{*}}(\theta_{*}) - J_{\pi_{\hat k}}(\hat \theta_{k})|| &\leq \gamma\beta||(\theta_{*} - \hat \theta_{k})J_{\pi_{*}}(\theta_{*})||
\end{aligned}
\end{equation}
Based on the Hoeffding’s inequality, we then bound $||(\theta_{*} - \hat \theta_{k})J_{\pi_{*}}(\theta_{*})||$ for all $s \in S$ in high probability.
\begin{equation}
Pr\left(||((\theta_{*} - \hat \theta_{k})J_{\pi_{*}}(\theta_{*}))(s)|| \geq \varepsilon\right) \leq 2 \exp \left(\frac{-K \varepsilon^{2}}{2 \beta^{2}}\right)
\end{equation}
By applying the union bound, we could then deduce:
\begin{equation}
\label{equation18}
Pr\left(||(\theta_{*} - \hat \theta_{k})J_{\pi_{*}}(\theta_{*})|| \geq \varepsilon\right) \leq 2 |S|\exp \left(\frac{-K \varepsilon^{2}}{2 \beta^{2}}\right)
\end{equation}
Define the probability of failure $\delta$ as:
\begin{equation}
\label{equation19}
\delta \triangleq 2|S| \exp \left(\frac{-K \epsilon^{2}}{2 \beta^{2}}\right)
\end{equation}
Define parameter $C$ as $C = \beta \sqrt{2 \log (2|S| / \delta) / K}$. By combing Equation \ref{equation18} and Equation \ref{equation19}, we could deduce:
\begin{equation}
\label{equation20}
Pr\left[||\left(\theta_{*} - \hat \theta_{k})J_{\pi_{*}}(\theta_{*})|| \geq \varepsilon\right)<C\right] \geq 1-\delta
\end{equation}
The bound for average regret could be deducted by combining Equation \ref{equation20} and Equation \ref{equation16}. Therefore, we could have:
\begin{equation}
||J_{\pi_{*}}(\theta_{*}) - \frac{1}{K}\sum_{k=1}^{K}r_{\pi_{\hat k}} (s_{k},a_{k})||
\leq C
\end{equation}
As for the total regret bound:
\begin{equation}
\begin{aligned}
KJ_{\pi_{*}}(\theta_{*}) - \sum_{k=1}^{K}r_{\pi_{\hat k}}(s_{k},a_{k}) &\leq KJ_{\pi_{*}}(\theta_{*}) - KJ_{\pi_{\hat k}}(\hat \theta_{k})\\
&\leq K||J_{\pi_{*}}(\theta_{*}) - KJ_{\pi_{\hat k}}(\hat \theta_{k})||\\
&\leq \beta \sqrt{2 K\log (2|S| / \delta)}
\end{aligned}
\end{equation}
Based on Lemma \ref{lemma1.2}, we could know that total number of the episodes is bounded by $K_{T} \leq \sqrt{2 S A T \log (T)}$. Therefore, we could bound $Reg_{K}$ by:
\begin{equation}
\begin{aligned}
Reg_{K} &\leq \beta \sqrt{2\sqrt{2 S A T \log (T)} \log (2|S| / \delta)}\\
&\leq \tilde{\mathcal{O}}(\sqrt{2\sqrt{SAT}})
\end{aligned}
\end{equation}
\end{proof}
\end{theorem}

\begin{theorem}
The regret for conducting the $\epsilon$-optimal policy could be bounded by:
$$
Reg_{\epsilon}(s) \leq 2 \gamma \beta \epsilon
$$
\begin{proof}
Assume there is a state $z$ that achieves the maximum regret. Define the optimal action as $a = \pi_{*}(z)$ and the near-optimal action as $b = \tilde{\pi}(z)$. Denote the optimal state value function at state $s$ as $J_{*}(s)$. The empirical state value function at state $s$ is defined as $\tilde{J}(s)$. Then We could have the inequality:
\begin{equation}
\label{equation23}
R(z, a)+\gamma \sum_{s \in S} \theta_{z,s}(a) \tilde{J}(s) \leq R(z, b)+\gamma \sum_{s \in S} \theta_{z,s}(b) \tilde{J}(s)
\end{equation}
Based on the previous assumption(Assumption \ref{assumption1}), we could have $J_{*}(s)-\epsilon \leq \tilde{J}(s) \leq J_{*}(s)+\epsilon$. Combining it with Equation \ref{equation23}, we could have
\begin{equation}
R(z, a)+\gamma \sum_{s \in S} \theta_{z,s}(a)\left(J_{*}(s)-\epsilon\right) \leq R(z, b)+\gamma \sum_{s \in S} \theta_{z,s}(b)\left(J_{*}(s)+\epsilon\right)
\end{equation}
Therefore, we could have:
\begin{equation}
R(z, a)-R(z, b) \leq 2 \gamma \epsilon+\gamma \sum_{s}\left[\theta_{z,s}(b) J_{*}(y)-\theta_{z,s}(a) J_{*}(y)\right]
\end{equation}
The maximum regret achieved on state $z$ could be defined as:
\begin{equation}
\begin{aligned}
Reg_{\epsilon}(z) &= J_{*}(z)-\tilde{J}(z) \\
&=R(z, a)-R(z, b)\\
&+\gamma \sum_{s}\left[\theta_{z,s}(a) J_{*}(s)-\theta_{z,s}(b) \tilde{J}(s)\right]
\end{aligned}
\end{equation}
Based on the previous assumption (Assumption \ref{assumption1}), we could have:
\begin{equation}
\begin{aligned}
Reg_{\epsilon}(z) & \leq 2\gamma \epsilon+\gamma \sum_{s}\left[\theta_{z,s}(b) J_{*}(s)-\theta_{z,s}(a) J_{*}(s)+\theta_{z,s}(a) J_{*}(s)-\theta_{z,s}(b) \tilde{J}(s)\right] \\
& \leq 2 \gamma \epsilon+\gamma \sum_{s} \theta_{z,s}(b)\left[J_{*}(s)-\tilde{J}(s)\right] \\
& \leq 2 \gamma \epsilon+\gamma \sum_{s} \theta_{z,s}(b) Reg_{\epsilon}(s)
\end{aligned}
\end{equation}
As we defined before, for all $s \in S$, $Reg_{\epsilon}(z) \geq Reg_{\epsilon}(s)$. We could deduce that:
\begin{equation}
Reg_{\epsilon}(z) \leq 2 \gamma \epsilon+\gamma \sum_{s} \theta_{z,s}(b) Reg_{\epsilon}(z)
\end{equation}
Therefore, the Lemma could be proved.
\begin{equation}
Reg_{\epsilon}(z) \leq \frac{2 \gamma \epsilon}{1-\gamma} = 2 \gamma \beta \epsilon
\end{equation}
\end{proof}
\end{theorem}

\begin{theorem}
The regret in time step $T$ could be bounded by:
$$
Reg_{T} \leq \tilde{\mathcal{O}}(\mathcal{D\sqrt{SAT}})
$$

\begin{proof}
We could know the definition of the regret in time step $T$ would be:
\begin{equation}
\begin{aligned}
Reg_{T} &= TJ_{\pi_{\hat{k}}}(\hat \theta) - \sum_{t = 1}^{T}r_{\pi_{t}}(s_{t},a_{t})\\
&= Reg_{T}^{1} + Reg_{T}^{2} + Reg_{T}^{3}
\end{aligned}
\end{equation}
We define the following regret as:
\begin{equation}
\begin{array}{l}
Reg_{T}^{1} = TJ_{\pi_{\hat{k}}}(\hat \theta) - \sum_{t = 1}^{T}J_{\pi_{t}}(\hat \theta) \\ \\
Reg_{T}^{2}=\sum_{t = 1}^{T}J_{\pi_{t}}(\hat \theta) - \sum_{t = 1}^{T}J_{\pi_{t}}(\theta_{t}) \\ \\
Reg_{T}^{3}= \sum_{t = 1}^{T}J_{\pi_{t}}(\theta_{t}) - \sum_{t = 1}^{T}r_{\pi_{t}}(s_{t},a_{t})
\end{array}
\end{equation}
Based on Theorem \ref{theorem4},Theorem \ref{theorem5},Theorem \ref{theorem6}, we could know that the bound of each regret are:
\begin{equation}
\begin{array}{l}
Reg_{T}^{1} = \tilde{\mathcal{O}} (\sqrt{T}/S^{2}) \\ \\
Reg_{T}^{2}=\tilde{\mathcal{O}} (D\sqrt{SAT}) \\ \\
Reg_{T}^{3}= \tilde{\mathcal{O}} (D\sqrt{ST})
\end{array}
\end{equation}
Therefore, we could deduce the bound of the total regret under time step $T$.
\end{proof}
\end{theorem}

\begin{lemma}
$$
\log \frac{J_{\pi_{*}}(\theta)}{J_{\pi_{t}}(\theta)} \leq \frac{\epsilon}{\Gamma}
$$
\begin{proof}
First, we could multiply $J_{\pi_{t}}(\theta)$ in order to construct the inequality. Let $J_{\pi_{t}}(\theta) = n$, $\epsilon = x$
\begin{equation}
\begin{aligned}
\lim _{n \rightarrow+\infty}\left(1+\frac{x}{n}\right)^{n} &= \lim _{n \rightarrow+\infty} e^{n \ln \left(1+\frac{x}{n}\right)} \\
&=e^{\lim _{n \rightarrow+\infty} \frac{\ln \left(1+\frac{x}{n}\right)}{\frac{1}{n}}}
\end{aligned}
\end{equation}
Apply the L'Hopital's Rule:
\begin{equation}
\begin{aligned}
\lim _{n \rightarrow+\infty}\left(1+\frac{x}{n}\right)^{n}&= e^{\lim _{n \rightarrow+\infty} \frac{\left(\frac{-x}{n^{2}}\right) \frac{1}{1+\frac{x}{n}}}{-\frac{1}{n^{2}}}}\\
&= e^{\lim _{n \rightarrow+\infty} \frac{x}{1+\frac{x}{n}}}=e^{x}
\end{aligned}
\end{equation}
Then, we could prove that $\left(1+\frac{x}{n}\right)^{n}$ is monotonically increased with $n$:
\begin{equation}
\begin{aligned}
(1+\frac{x}{n})^{2} &= 1 \cdot\underbrace{ \left(1+\frac{x}{n}\right) \cdot\left(1+\frac{x}{n}\right) \cdots \cdots\left(1+\frac{x}{n}\right)}_{n}\\
& \leq \left[\frac{1+(1+\frac{x}{n})+ \cdots +(1+\frac{x}{n})}{n+1}\right]^{n+1}\\
& = \left[\frac{1+n(1+\frac{x}{n})}{n+1}\right]^{n+1}\\
& = \left[1+\frac{x}{n(n+1)}\right]^{n+1}\\
& \leq \left[1+\frac{x}{n+1}\right]^{n+1}
\end{aligned}
\end{equation}
The first inequality holds for the arithmetic mean equality. We could deduce that $(1+\frac{x}{n})^{n} \leq e^{x}$. Therefore, we have:
\begin{equation}
J_{\pi_{t}}(\theta) \log \frac{J_{\pi_{*}}(\theta)}{J_{\pi_{t}}(\theta)} \leq \epsilon
\end{equation}
Based on Assumption \ref{assumption4}, we could deduce the upper bound of average discounted reward. Then the lemma could be proved.
\end{proof}
\end{lemma}

\begin{lemma}
\label{lemma3replacement}
Under Assumption \ref{assumption3}, for each stationary near-optimal policy $\pi$ and epoch counter $k \geq 1$. The following upper bound holds for negative log-density.
$$
-\log W_{t_{k}}(\pi) \leq \frac{\epsilon}{\Gamma}|S|^{2}(  \rho(k_{\pi}) \sqrt{k_{\pi}} + k_{\pi} \tilde{\tau}_{t_{k},k_{\pi}})
$$
\begin{proof}
When $W_{t_{k}} \leq 1$, we could have:
\begin{equation}
W_{t_{k}}(\theta):=\exp \sum_{\pi, s_{1}, s_{2}} \mathcal{H}\left(N_{\pi}(k), \pi\right) \log \frac{J_{\pi_{t}}(\theta)}{J_{\pi_{*}}(\theta)}
\end{equation}
Based on the definition of the counter $\mathcal{H}$, we could deduce the value of the posterior weight in a single epoch:
\begin{equation}
\begin{aligned}
W_{t_{k}}(\theta)&= \exp \left(\sum_{t=1}^{\infty} \mathbbm{1}\left\{\pi_{e(t)}=\pi,\left(S_{t}, S_{t+1}\right)=\left(s_{1}, s_{2}\right), N(e(t)) \leq k\right\}\log \frac{J_{\pi_{t}}(\theta)}{J_{\pi_{*}}(\theta)}\right)\\
& = \exp \left(\sum_{\pi \in \Pi} \sum_{\left(s_{1}, s_{2}\right) \in \mathcal{S}^{2}} \sum_{t=1}^{T} \mathbbm{1}\left\{\pi_{e(i)}=\pi,\left(S_{t}, S_{t+1}\right)=\left(s_{1}, s_{2}\right)\right\}\log \frac{J_{\pi_{t}}(\theta)}{J_{\pi_{*}}(\theta)}\right)\\
& = \exp \left(N_{\pi}(t) \sum_{\left(s_{1}, s_{2}\right) \in \mathcal{S}^{2}} \sum_{t=0}^{t-1} \frac{\mathbbm{1}\left\{\pi_{e(t)}=\pi,\left(S_{t}, S_{t+1}\right)=\left(s_{1}, s_{2}\right)\right\}}{N_{\pi}(t)}\log \frac{J_{\pi_{t}}(\theta)}{J_{\pi_{*}}(\theta)}\right)\\
\end{aligned}
\end{equation}
Where $N_{\pi}(t):=\sum_{t=0}^{t-1}\sum_{\pi \in \Pi} \mathbbm{1}\left\{\pi_{e(t)}=\pi\right\}$ represents the total number of the time instants during the period of $t$ when policy $\pi$ was conducted.

When Assumption \ref{assumption3} holds, we could know that $N_{\pi}(t) = \tilde{\tau}_{\pi_{t_{k}},N_{\pi}(k)}$, where $N_{\pi}(k) :=\sum_{k=0}^{K}\sum_{\pi \in \Pi} \mathbbm{1}\left\{\pi_{e(k)}=\pi\right\} $ holds for the number of the epochs where policy $\pi$ was chosen(The notation of $\tau$ will be represented as $N_{\pi}(k) = k_{\pi}$, $\tilde{\tau}_{\pi_{t_{k}},N_{\pi}(k)} = \tilde{\tau}_{t_{k},k_{\pi}}$). Therefore, we could have:
\begin{equation}
\begin{aligned}
& -\log W_{t_{k}}(\pi)\\
& = -N_{\pi}(t) \sum_{\left(s_{1}, s_{2}\right) \in \mathcal{S}^{2}} \sum_{t=0}^{t-1} \frac{\mathbbm{1}\left\{\pi_{e(t)}=\pi,\left(S_{t}, S_{t+1}\right)=\left(s_{1}, s_{2}\right)\right\}}{N_{\pi}(t)}\log \frac{J_{\pi_{t}}(\theta)}{J_{\pi_{*}}(\theta)}\\
& = -\sum_{\left(s_{1}, s_{2}\right) \in \mathcal{S}^{2}} \tilde{\tau}_{t_{k},k_{\pi}}  \mathcal{H}_{\left(s_{1}, s_{2}\right)}\left(\tilde{\tau}_{t_{k},k_{\pi}}, \pi\right) \log \frac{J_{\pi_{t}}(\theta)}{J_{\pi_{*}}(\theta)}\\
& = \sum_{\left(s_{1}, s_{2}\right) \in \mathcal{S}^{2}} \left[\tilde{\tau}_{t_{k},k_{\pi}}  \mathcal{H}_{\left(s_{1}, s_{2}\right)}\left(\tilde{\tau}_{t_{k},k_{\pi}}, \pi\right) - k_{\pi}\tilde{\tau}_{t_{k},k_{\pi}}\theta_{\pi}(s_{1}|s_{2})\right]\log \frac{J_{\pi_{*}}(\theta)}{J_{\pi_{t}}(\theta)} + \sum_{\left(s_{1}, s_{2}\right) \in \mathcal{S}^{2}} k_{\pi}\tilde{\tau}_{t_{k},k_{\pi}}\theta(s_{1}|s_{2}) \log \frac{J_{\pi_{*}}(\theta)}{J_{\pi_{t}}(\theta)}\\
\end{aligned}
\end{equation}
The last equation is based on the logarithmic property $\log \frac{A}{B} = - \log \frac{B}{A}$. Based on the Assumption \ref{assumption3}, define $\rho(x) := O(\sqrt{\log \log(x)})$.
\begin{equation}
\begin{aligned}
- \log W_{t_{k}}(\pi)& \leq \sum_{\left(s_{1}, s_{2}\right) \in \mathcal{S}^{2}} \rho(k_{\pi}) \sqrt{k_{\pi}} \log \frac{J_{\pi_{*}}(\theta)}{J_{\pi_{t}}(\theta)}+ k_{\pi}\tilde{\tau}_{t_{k},k_{\pi}} \sum_{\left(s_{1}, s_{2}\right) \in \mathcal{S}^{2}}\theta(s_{1}|s_{2}) \log \frac{J_{\pi_{*}}(\theta)}{J_{\pi_{t}}(\theta)}\\
& \leq \frac{\epsilon}{\Gamma}|S|^{2}(  \rho(k_{\pi}) \sqrt{k_{\pi}} + k_{\pi} \tilde{\tau}_{t_{k},k_{\pi}})
\end{aligned}
\end{equation}
\end{proof}
\end{lemma}

\begin{lemma}
\label{lemma3.3replacement}
The difference between the local optimal average reward and the instantaneous average reward could be bounded by:
$$
|J_{\pi_{t}}-J^{*}| \leq \tilde{\mathcal{O}}(\frac{1}{\sqrt{T}})
$$
\begin{proof}
We could know that the current policy probability distribution is updated based on the previous distribution and the current optimal policy distribution:
\begin{equation}
\mu_{t}(\pi) = W_{t}\mu_{t}(\pi) + (1-W_{t})\pi_{t}^{*}(s,\theta_{t_{k}}
\end{equation}
We could extend this result to reward function:
\begin{equation}
\label{equation50}
\begin{aligned}
J_{\pi_{t}}&=W_{t} J_{\pi_{t-1}}+\left(1-W_{t}\right) J^{*}\left(\theta_{t}\right)\\
J_{\pi_{t}}^{2} &=W_{t}^{2} J_{\pi{t-1}}^{2}+\left(1-W_{t}\right)^{2} J^{*^{2}}+2 W_{t}\left(1-W_{t}\right) J^{*} J_{\pi_{t-1}} \\
& \leq W_{t}^{2} J_{\pi t}^{2}+\left(1-W_{t}\right)^{2} J^{* 2}+2 W_{t}\left(1-W_{t}\right) J^{*} J_{\pi_{t}}
\end{aligned}
\end{equation}
The inequality is based on the monotonicity of the algorithm. We could simplify Equation \ref{equation50}:
\begin{equation}
\begin{aligned}
\left(1-W_{t}^{2}\right) J_{\pi_{t}}^{2} &\leq \left(1-W_{t}\right)^{2} J^{*^{2}}+2 W_{t}\left(1-W_{t}\right) J^{*} J_{\pi_{t}} \\
\left(1+W_{t}\right) J_{\pi t}^{2} &\leq \left(1-W_{t}\right) J^{*^{2}}+2 W_{t} J^{*} J_{\pi_{t}} \\
J_{\pi_{t}}^{2}+W_{t} J_{\pi_{t}}^{2} &\leq J^{*^{2}}-W_{t} J^{*^{2}}+2 W_{t} J^{*} J_{\pi_{t}}\\
W_{t}\left(J_{\pi_{t}}^{2}+J^{*^{2}}\right) &\leq J^{*^{2}}-J_{\pi_{t}}^{2}+2 W_{t} J^{*} J_{\pi_{t}}\\
J_{\pi_{t}}^{2}+J^{*^{2}} &\leq \frac{1}{W_{t}}\left(J^{*^{2}}-J_{\pi_{t}}^{2}\right)+2 J^{*} J_{\pi_{t}}
\end{aligned}
\end{equation}
Based on the definition of the regret of each time step, we could deduce the bound of the instantaneous regret:
\begin{equation}
\begin{aligned}
\left(J_{\pi t}-J^{*}\right)^{2} &=J_{\pi_{t}}^{2}+J^{*^{2}}-2 J_{\pi t} J^{*} \\
& \leq \frac{1}{w_{t}}\left(J^{*^{2}}-J_{\pi_{t}}^{2}\right)+2 J^{*} J_{\pi_{t}}-2 J_{\pi_{t}} J^{*} \\
&=\frac{1}{w_{t}}\left(J^{*^{2}}-J_{\pi_{t}}^{2}\right) \\
&=\frac{1}{w_{t}}\left(J^{*}-J_{\pi_{t}}\right)\left(J^{*}+J_{\pi_{t}}\right)
\end{aligned}
\end{equation}

\begin{equation}
|J_{\pi_{t}}-J^{*}| \leq \frac{1}{W_{t}}|J_{\pi_{t}}+J^{*}|
\end{equation}
From Lemma \ref{lemma3}, we could know that $-\log W_{t_{k}}(\pi)$ is bounded by $B$, with $B = \frac{\epsilon}{\Gamma}|S|^{2}(\rho(k_{\pi}) \sqrt{k_{\pi}} + k_{\pi} \tilde{\tau}_{t_{k},k_{\pi}})$. Therefore, we could construct the following inequalities.
\begin{equation}
\begin{aligned}
W_{t_{k}}-1 &\geq \log W_{t_{k}} \geq -B\\
\frac{1}{W_{t_{k}}} &\leq \frac{1}{1-B}
\end{aligned}
\end{equation}
Factor $B$ is proportional to parameter $k_{\pi}$ which could be bounded by the total number of episode of under total time $T$. Therefore, we could bound $\frac{1}{W_{t}}$ by $T$(Ignoring the constants):
\begin{equation}
\begin{aligned}
\frac{1}{W_{t}} &\leq \frac{1}{1-\frac{\epsilon}{\Gamma}|S|^{2}(\rho(k_{\pi}) \sqrt{k_{\pi}} + k_{\pi} \tilde{\tau}_{t_{k},k_{\pi}})}\\
&\leq \frac{1}{1-\sqrt{\sqrt{T}}-\sqrt{T}}
\end{aligned}
\end{equation}
Based on Assumption \ref{assumption4}, the average discounted reward function is bounded by $\Gamma$. So the difference between the local optimal average reward and the instantaneous average reward could be bounded by:
\begin{equation}
\begin{aligned}
|J_{\pi_{t}}-J^{*}| &\leq \frac{1}{W_{t}}|J^{*}+J_{\pi_{t}|}\\
&\leq \frac{2}{1-\frac{\epsilon}{\Gamma}|S|^{2\Gamma}(\rho(k_{\pi}) \sqrt{k_{\pi}} + k_{\pi} \tilde{\tau}_{t_{k},k_{\pi}})}\\
&\leq \tilde{\mathcal{O}} (\frac{2\Gamma^{2}}{S^{2}\sqrt{T}})
\end{aligned}
\end{equation}
\end{proof}
\end{lemma}

\begin{theorem}
\label{theorem4}
The first part of the regret in time step $T$ is bounded by:
$$
Reg_{T}^{1} \leq \tilde{\mathcal{O}}(\frac{\sqrt{T}}{S^{2}})
$$
\begin{proof}
From the definition before, we could know that $Reg_{T}^{1}$ could be represented as:
\begin{equation}
Reg_{T}^{1} = TJ_{\pi_{\hat{k}}}(\hat \theta) - \sum_{t = 1}^{T}J_{\pi_{t}}(\hat \theta)
\end{equation}
Since this theorem won't involve the transformation of the transition probability. So let $J_{\pi}(\theta) = J_{\pi}$. Based on the update rule of the posterior distribution $\mu_{t+1}(\pi)$ of policy $\pi$. We could divide the average discounted return into several parts:\\
At time step $t = T$, we could assume the instantaneous regret equals to zero:
\begin{equation}
Reg_{t_{T}}^{1} = J_{\pi_{\hat k}}- J_{\pi_{T}} = 0
\end{equation}
At time step $t = T-1$, define the local optimal average discounted return as $J_{\pi}^{*}$. Note that this local optimal value is virtual. The instantaneous regret could be represented as:
\begin{equation}
\begin{aligned}
Reg_{t_{T-1}}^{1} &= J_{\pi_{\hat k}} - J_{\pi_{T-1}}\\
&= W_{t-1}J_{\pi_{T-1}} + (1-W_{t-1})J_{\pi}^{*} - J_{\pi_{T-1}}\\
&= (W_{t-1}-1)J_{\pi_{T-1}} + (1-W_{t-1})J_{\pi}^{*}\\
&= (1-W_{t-1})(J_{\pi}^{*} - J_{\pi_{T-1}})
\end{aligned}
\end{equation}
In a similar fashion, at time step $t = T-2$, the instantaneous regret could be represented as:
\begin{equation}
\begin{aligned}
Reg_{t_{T-2}}^{1} &= J_{\pi_{\hat k}} - J_{\pi_{T-2}}\\
&= J_{\pi_{\hat k}} - J_{\pi_{T-1}} + J_{\pi_{T-1}} - J_{\pi_{T-2}}\\
&= (1-W_{t-1})(J_{\pi}^{*} - J_{\pi_{T-1}}) + (1-W_{t-2})(J_{\pi}^{*} - J_{\pi_{T-1}})
\end{aligned}
\end{equation}
Based on Lemma \ref{lemma3.3}, the difference between the local optimal value and the current average return could be bounded by:
\begin{equation}
|J_{\pi}^{*} - J_{\pi_{t}}| \leq \tilde{\mathcal{O}}(\frac{1}{\sqrt{T}})
\end{equation}

The sub-optimal models are sampled when their posterior probability is larger than $\frac{1}{T}$. This ensures the time complexity of the Thompson sampling process is no more than $O(1)$. So we could deduce the total regret in time step $T$.
\begin{equation}
\begin{aligned}
Reg_{T}^{1} &= \frac{1}{T} (Reg_{t_{T-1}}^{1} + Reg_{t_{T-2}}^{1} + \cdots + Reg_{t_{1}}^{1})\\
&\leq \tilde{\mathcal{O}}(\frac{2\Gamma^{2}}{S^{2}\sqrt{T}})(\frac{T-1}{T} + \frac{T-2}{T} + \cdots \frac{1}{T})\\
&\leq \tilde{\mathcal{O}}(\frac{\Gamma^{2}\sqrt{T}}{S^{2}})
\end{aligned}
\end{equation}
\end{proof}
\end{theorem}

\begin{theorem}
\label{theorem5}
The second part of the regret in time step $T$ is bounded by:
$$
Reg_{T}^{2} \leq \tilde{\mathcal{O}}( D\sqrt{SAT} )
$$
\begin{proof}
From the definition before, we could know that $Reg_{T}^{2}$ could be represented as:
\begin{equation}
Reg_{T}^{2}=\sum_{t=1}^{T} J_{\pi_{t}}(\hat{\theta})-\sum_{t=1}^{T} J_{\pi_{t}}\left(\theta_{t}\right)
\end{equation}
%Based on the update rule of the transition probability, the posterior weight of the transition probability could be represented as the likelihood ratio of the history MDP:
%\begin{equation}
%\begin{aligned}
%&W_{t}(\theta) :=\prod_{t=0}^{T} \frac{\mu_{t}(\theta)\left(S_{t}, A_{t}, S_{t+1}\right)}{\mu_{t}(\theta^{\prime})\left(S_{t}, A_{t}, S_{t+1}\right)} \\
%&=\exp \left(\sum_{\pi \in \Pi} \sum_{t=0}^{T} \mathbbm{1}\left\{\pi_{e(t)}=\pi\right\} \log  \frac{\mu_{t}(\theta)\left(S_{t}, A_{t}, S_{t+1}\right)}{\mu_{t}(\theta^{\prime})\left(S_{t}, A_{t}, S_{t+1}\right)}\right)\\
%&= \exp \left(\sum_{\pi \in \Pi} \sum_{\left(s_{1}, s_{2}\right) \in \mathcal{S}^{2}} \sum_{t=1}^{T} \mathbbm{1}\left\{\pi_{e(i)}=\pi,\left(S_{t}, S_{t+1}\right)=\left(s_{1}, s_{2}\right)\right\}\right.\\
%&\left.\quad \log  \frac{\mu_{t}(\theta)\left(S_{t}, A_{t}, S_{t+1}\right)}{\mu_{t}(\theta^{\prime})\left(S_{t}, A_{t}, S_{t+1}\right)}\right)\\
%& =\exp \left(-\sum_{\pi \in \Pi} N_{\pi}(t) \sum_{\left(s_{1}, s_{2}\right) \in \mathcal{S}^{2}} \sum_{t=0}^{T}\right.\\
%&\left.\quad \frac{\mathbbm{1}\left\{\pi_{e(t)}=\pi,\left(S_{t}, S_{t+1}\right)=\left(s_{1}, s_{2}\right)\right\}}{N_{\pi}(t)} \log  \frac{\mu_{t}(\theta)\left(S_{t}, A_{t}, S_{t+1}\right)}{\mu_{t}(\theta^{\prime})\left(S_{t}, A_{t}, S_{t+1}\right)}\right)
%\end{aligned}
%\end{equation}
In this theorem, we mainly focus on the difference between transition probability. So we could denote the average discounted return as $J_{\pi_{t}}(\hat \theta) = J(\hat \theta)$, $J_{\pi_{t}}(\theta_{t}) = J(\theta_{t})$. Note that the $\sum_{t=1}^{T}J(\hat \theta)$ is a virtual value which represents the average discounted return in time step $T$ with transition probability $\hat \theta$. From the previous definition of the Bellman iterator of the average discounted return, we could have:

%Since the transition probability $\hat \theta$ and $\theta_{t}$ are identically distributed on the history. That means the  for any measurable function $f$ and any $h_{t_{k}}$-measurable random variable $X$, we have $\mathbb{E}\left[f\left(\hat \theta, X\right) \mid h_{t_{k}}\right]=\mathbb{E}\left[f\left(\theta_{k}, X\right) \mid h_{t_{k}}\right]$. Therefore, we could deduce the relationship between average discounted return under optimal transition probability and near-optimal transition probability.
\begin{equation}
\begin{aligned}
&J(\hat \theta) + b(\hat \theta, \pi, s) = r(s,\pi) + \sum_{s^{\prime}}\hat \theta(s^{\prime}|s,\pi)b(\hat \theta, \pi, s^{\prime})\\
&J(\theta_{t}) + b(\theta_{t}, \pi, s) = r(s,\pi) + \sum_{s^{\prime}}\theta_{t}(s^{\prime}|s,\pi)b(\theta_{t}, \pi, s^{\prime})\\
\end{aligned}
\end{equation}
For brevity, let $b(\hat \theta,\pi,s) = b(\hat \theta)$. The difference between the average discounted return under optimal transition probability and near-optimal transition probability could be represented as:
\begin{equation}
\begin{aligned}
J(\hat \theta) - J(\theta_{t}) &= b(\theta_{t}, \pi, s) - b(\hat \theta, \pi, s) \\
&+ \sum_{s^{\prime}}\hat \theta(s^{\prime}|s,\pi)b(\hat \theta, \pi, s^{\prime}) - \theta_{t}(s^{\prime}|s,\pi)b(\theta_{t}, \pi, s^{\prime})\\
\end{aligned}
\end{equation}
We could bound the first term with the largest difference between each state:
\begin{equation}
\label{equation55}
\begin{aligned}
0 \leq b(\hat \theta, \pi, s) - b(\theta_{t}, \pi, s) &\leq sp(b(\theta)) \leq D\\
0 \geq b(\theta_{t},\pi,s) - b(\hat \theta, \pi, s) & \geq -D
\end{aligned}
\end{equation}
%Since the transition probability $\hat \theta$ and $\theta_{t}$ are identically distributed on the history. That means the  for any measurable function $f$ and any $h_{t_{k}}$-measurable random variable $X$, we have $\mathbb{E}\left[f\left(\hat \theta, X\right) \mid h_{t_{k}}\right]=\mathbb{E}\left[f\left(\theta_{k}, X\right) \mid h_{t_{k}}\right]$.
Based on Equation \ref{equation55}, we could bound the second term in a similar way:
\begin{equation}
\begin{aligned}
&\sum_{s^{\prime}}\hat \theta(s^{\prime}|s,\pi)b(\hat \theta, \pi, s^{\prime}) - \theta_{t}(s^{\prime}|s,\pi)b(\theta_{t}, \pi, s^{\prime})\\
&\leq D\sum_{s^{\prime}}\left(\hat \theta(s^{\prime}|s,\pi) - \theta_{t}(s^{\prime}|s,\pi)\right)\\
\end{aligned}
\end{equation}
Based on the previous work(Weissman et al2003), we could bound the difference between the near-optimal transition probability and the instantaneous probability with confidence set $C_{t}$. The confidence set $C_{t}$ is defined as:
\begin{equation}
C_{t} := \left\{\theta: \sum_{s^{\prime}}\left|\theta\left(s^{\prime} \mid s, a\right)-\hat{\theta}\left(s^{\prime} \mid s, a\right)\right| \leq b_{t}(s, a) \quad \forall s, a, s^{\prime}\right\}
\end{equation}
where $b_{t}(s, a):=\sqrt{\frac{14 S \log \left(2 A t T\right)}{\max \left\{1, N_{t_{k}}(s, a)\right\}}}, t \leq T$. The counter function is defined as $N_{t}(s,a) = \sum_{t \in T}\mathbbm{1}{s_{t} = s,a_{t} = a }$. Therefore, we could bound the denominator of $b_{t}$ in time step $T$:
\begin{equation}
\begin{aligned}
\sum_{t=1}^{T} \sqrt{\frac{1}{\max \left\{1, N_{t}\left(s_{t}, a_{t}\right)\right\}}}&=\sum_{t=1}^{T} \sum_{s, a} \frac{\mathbbm{1}\left\{s_{t}=s, a_{t}=a\right\}}{\sqrt{\max \left\{1, N_{t}(s, a)\right\}}}\\
&=\sum_{s, a} \sum_{t=1}^{T} \frac{\mathbbm{1}\left\{s_{t}=s, a_{t}=a\right\}}{\sqrt{\max \left\{1, N_{t}(s, a)\right\}}}\\
&=\sum_{s, a}\left(1+\sum_{j=1}^{n_{T+1}(s, a)-1} \frac{1}{\sqrt{j}}\right) \\
&\leq \sum_{s, a}\left(1+2 \sqrt{N_{T+1}(s, a)}\right)\\
&=S A+2 \sum_{s, a} \sqrt{N_{T+1}(s, a)} \\
&\leq S A+2 \sqrt{S A \sum_{s, a} N_{T+1}(s, a)}=S A+2 \sqrt{S A T}
\end{aligned}
\end{equation}
Therefore, we could bound $Reg_{T}^{2}$:
\begin{equation}
\begin{aligned}
Reg_{T}^{2} &= \sum_{t=1}^{T} J_{\pi_{t}}(\hat{\theta})-\sum_{t=1}^{T} J_{\pi_{t}}\left(\theta_{t}\right)\\
&=  \sum_{t=1}^{T} J(\hat \theta) - J(\theta_{t})\\
&\leq \sum_{t=1}^{T} \left(-D + D \sum_{s^{\prime}}\left(\hat{\theta}\left(s^{\prime} \mid s, \pi\right)-\theta_{t}\left(s^{\prime} \mid s, \pi\right)\right)\right)\\
&\leq  D\sum_{t=1}^{T} b_{t}(s,a)\\
&=  D\sum_{t=1}^{T} \sqrt{\frac{14 S \log \left(2 A t T\right)}{\max \left\{1, N_{t_{k}}(s, a)\right\}}}\\
&\leq  D\sum_{t=1}^{T}\sqrt{\frac{14 S \log \left(2 A T^{2}\right)}{\max \left\{1, N_{t_{k}}(s, a)\right\}}}\\
&\leq  D\sqrt{28 S \log (2 A T)}(S A+2 \sqrt{S A T})
\end{aligned}
\end{equation}
Therefore, we could finish the proof of the bound of the $Reg_{T}^{2}$

\end{proof}
\end{theorem}

\begin{theorem}
\label{theorem6}
The third part of the regret in time step $T$ is bounded by:
$$
Reg_{T}^{3} \leq \tilde{\mathcal{O}}(D\sqrt{ST})
$$
\begin{proof}
From the definition before, we could know that $Reg_{T}^{3}$ could be represented as:
\begin{equation}
\label{equation60}
Reg_{T}^{3}=\sum_{t=1}^{T} J_{\pi_{t}}\left(\theta_{t}\right)-\sum_{t=1}^{T} r_{\pi_{t}}\left(s_{t},\theta_{t}\right)
\end{equation}
We introduce the intermediate variable $J_{\bar \pi}(\theta_{t})$ to represent the average discounted reward under average policy $\bar \pi$ at time step $t$. Note that this variable is virtual and can not be spotted in the reality.
\begin{equation}
\begin{aligned}
Reg_{T}^{3}&=\sum_{t=1}^{T} J_{\pi_{t}}\left(\theta_{t}\right)- \sum_{t=1}^{T} J_{\bar \pi}\left(\theta_{t}\right)+\sum_{t=1}^{T} J_{\bar \pi}\left(\theta_{t}\right)-\sum_{t=1}^{T} r_{\pi_{t}}\left(s_{t}, \theta_{t}\right)\\
&= \sum_{t=1}^{T}Y_{t}^{1}+\sum_{t=1}^{T}Y_{t}^{2}
\end{aligned}
\end{equation}
Let $Y_{t}^{1}= J_{\pi_{t}}\left(\theta_{t}\right)- J_{\tilde{\pi}}\left(\theta_{t}\right)$,$Y_{t}^{2}= J_{\pi}\left(\theta_{t}\right)- r_{\pi_{t}}\left(s_{t}, \theta_{t}\right)$,we then bound these two variables using its Bellman iterator form. Denote the total reach time of state $s$ at time step $t$ as $N_{s}(t) = \sum_{t=1}^{T}\mathbbm{1}_{s_{t} = s}$. By the construction of $\bar \pi$, we could deduct the average form of the transition probability. Note that the average values here are all virtual:

%From the upper Equation \ref{equation61} and the monotonicity of our algorithm. The inequality holds for:
%\begin{equation}
%\label{equation62}
%\begin{aligned}
%J\left(\theta_{t}\right)-r(s, \pi_{t})&\leq \max \sum_{s^{\prime}} \theta_{t}\left(s^{\prime} \mid s, \pi\right) b\left(\theta_{t}, \pi, s^{\prime}\right)- b\left(\theta_{t}, \pi, s\right)\\
%&\leq \sum_{s^{\prime}}\hat \theta (s^{\prime}|s,\pi)b(\hat \theta, \pi, s^{\prime}) - b(\theta_{t},\pi,s)\\
%\end{aligned}
%\end{equation}
%We could know that the average form of the transition probability and the reward function could be represented as:
\begin{equation}
\label{equation63}
\begin{aligned}
\bar \theta\left(s^{\prime} \mid s, \pi\right)&=\sum_{a^{2}} \frac{\sum_{t=1}^{T} \mathbbm{1}_{s_{t}=s} \pi_{t}(a)}{N_{s}(t)} \theta \left(s^{\prime} \mid s, \pi_{t}, a\right)\\
&=\frac{1}{N_{s}(t)} \sum_{t=1}^{T} \mathbbm{1}_{s_{t}=s} \theta\left(s^{\prime} \mid s, \pi_{t}\right)\\
\end{aligned}
\end{equation}
In a similar fashion, we could deduce the average discounted reward function:
\begin{equation}
\label{equation64}
\begin{aligned}
J_{\bar \pi}(\theta_{t}) &= \frac{1}{\gamma} \bar r(s,\theta_{t}) = \sum_{a^{2}} \frac{\sum_{t=1}^{T} \mathbbm{1}_{s_{t}=s} \pi_{t}\left(a\right)}{\gamma N_{s}(t)} r\left(s, \pi_{t}, a,\theta_{t}\right)\\
&= \frac{1}{\gamma N_{s}(t)} \sum_{t=1}^{T} \mathbbm{1}_{s_{t}=s} r\left(s , \pi_{t},\theta_{t}\right)
\end{aligned}
\end{equation}
Based on the definition of the Bellman iterator, we could have:
\begin{equation}
\label{equation61}
\begin{aligned}
J\left(\theta_{t}\right)+b\left(\theta_{t}, \pi, s\right)&=r_{\pi_{t}}(s, \theta_{t})+\sum_{s^{\prime}} \theta_{t}\left(s^{\prime} \mid s, \pi\right) b\left(\theta_{t}, \pi, s^{\prime}\right)\\
J\left(\theta_{t}\right)-r_{\pi_{t}}(s, \theta_{t})&=\sum_{s^{\prime}} \theta_{t}\left(s^{\prime} \mid s, \pi\right) b\left(\theta_{t}, \pi, s^{\prime}\right)-b\left(\theta_{t}, \pi, s\right)\\
\end{aligned}
\end{equation}
Since the  we could get:
\begin{equation}
\begin{aligned}
\sum_{t=1}^{T}\left(J_{ \pi_{t}}(\theta_{t})-J_{\bar \pi}\left(\theta_{t}\right)\right)&\leq \sum_{t=1}^{T}\left(J_{ \pi_{t}}(\theta_{t})-{r}_{\pi_{t}}\left(s_{t},\theta_{t}\right)\right)
\\
&=\sum_{t=1}^{T}\left(\sum_{s^{\prime}} \theta_{t}\left(s^{\prime} \mid s_{t}\right) b\left(\theta_{t},s^{\prime}\right)-b\left(\theta_{t},s^{\prime}\right)\right) \\
&=\sum_{s, s^{\prime}} N_{s}(t) \theta_{t}\left(s^{\prime} \mid s\right)b\left(\theta_{t},s^{\prime}\right)-\sum_{t=1}^{T} b\left(\theta_{t},s_{t}\right) \\
&=\sum_{s, s^{\prime}} \sum_{t=1}^{T} \mathbbm{1}_{s_{t}=s} \theta\left(s^{\prime} \mid s, \pi_{t}\right) b\left(\theta_{t},s^{\prime}\right)-\sum_{t=1}^{T} b\left(\theta_{t},s_{t}\right) \\
&=\sum_{t=1}^{T} \sum_{s^{\prime}} \theta\left(s^{\prime} \mid s_{t}, \pi_{t}\right) b\left(\theta_{t},s^{\prime}\right)-\sum_{t=1}^{T} b\left(\theta_{t},s_{t}\right)
\end{aligned}
\end{equation}
We could spot that the virtual average discounted reward function $J_{\bar \pi}(\theta_{t})$ and the instantaneous reward $r(s_{t},\theta_{t})$ shares the same transition probability. Therefore, we could ignore the influence of the transition probability. For brevity, we denote the instantaneous reward $r_{\pi_{t}}(s_{t},\theta_{t})$ as $r_{t}$. From the deduction before, we could transform $Y_{t}^{2}$ into:
\begin{equation}
\begin{aligned}
\sum_{t=1}^{T}\left(J_{\bar \pi}(\theta_{t})-{r}_{\pi_{t}}\left(s_{t},\theta_{t}\right)\right)&= \sum_{s} \frac{N_{s}(t)}{\gamma} \bar{r}(s)-\sum_{t=1}^{T} r_{t}\\
&= \sum_{s} \sum_{t=1}^{T} \mathbbm{1}_{s_{t}=s} \frac{r\left(s,\pi_{t}\right)}{\gamma}-\sum_{t=1}^{T} r_{t}\\
&= \sum_{t=1}^{T} \frac{r(s,\pi_{t})}{\gamma} - \sum_{t=1}^{T}r_{t} \\
&\leq \sum_{t=1}^{T} r(s,\pi_{t}) - \sum_{t=1}^{T}r_{t}\\
&\leq 0
\end{aligned}
\end{equation}

Since the value of $Y_{t}^{2}$ is less than zero. The value of the total regret value should be directly relate to the bound of $Y_{t}^{1}$. From Lemma \ref{lemma6}, we could know that $Y_{t}^{1}$ is a martingale difference sequence. for every $b^{j} \ in D$, $j = 1,\cdots,(2DST)$. We could apply the Azuma-Hoeffding's inequality and bound $Y_{t}^{1}$:
\begin{equation}
\label{equation67}
\sum_{t=1}^{T} Y_{t}^{1,j} \leq \sqrt{\frac{\log \left((2 D S T)^{S} \delta^{-1}\right)}{2} T(2 D)^{2}}
\end{equation}
with probability at least $1-\frac{\delta}{2DST}$. Using the union bound, Equation \ref{equation67} holds for all $j$ with probability at least $1-\delta$. Thus, we could have:
\begin{equation}
\sum_{t=1}^{T} Y_{t}^{1} \leq \tilde{\mathcal{O}}(D \sqrt{S T})
\end{equation}

\end{proof}
\end{theorem}

\begin{lemma}
\label{lemma6}
To prove that $Y_{t}^{1}$ is a martingale difference sequence.
\begin{proof}
From the previous definition of $Y_{t}^{1,j}$, we could know that:
\begin{equation}
Y_{t}^{1,j} = \sum_{t=1}^{T} \sum_{s^{\prime}} \theta\left(s^{\prime} \mid s_{t}, \pi_{t}\right) b^{j}\left(\theta_{t}, s^{\prime}\right)-\sum_{t=1}^{T} b^{j}\left(\theta_{t}, s_{t}\right)
\end{equation}
The bias vector $b$ is $F_{t_{k}}$-measurable, where $\mathcal{F}_{t-1}:=\left\{s_{1}, a_{1}, \cdots, s_{t}\right\}$.
Therefore, the expectation value of $Y_{t}^{1,j}$ should be equals to zero.
\begin{equation}
\begin{aligned}
E[Y_{t}^{1,j}] &= E\left[ \sum_{s^{\prime}} \theta\left(s^{\prime} \mid s_{t}, \pi_{t}\right) b^{j}\left(\theta_{t},s^{\prime}\right)- b^{j}\left(\theta_{t},s_{t}\right)\right]\\
&= E\left[\sum_{s^{\prime}} \theta\left(s^{\prime} \mid s_{t}, \pi_{t}\right) b^{j}\left(\theta_{t},s^{\prime}\right)\right]- E\left[b^{j}\left(\theta_{t},s_{t}\right)\right]\\
&= 0
\end{aligned}
\end{equation}
Since the expectation value of $Y_{t}^{1,j}$ equals to zero, we could prove that $Y_{t}^{1} = \sum_{j}Y_{t}^{1,j}$ is a martingale difference sequence.
\end{proof}
\end{lemma}
%%%%%%%%%%%%%%%%%%%%%%%%%%%%%%%%%%%%%%%%%%%%%%%%%%%%%%%%%%%%%%%%%%%%%%%%%%%%%%%
%%%%%%%%%%%%%%%%%%%%%%%%%%%%%%%%%%%%%%%%%%%%%%%%%%%%%%%%%%%%%%%%%%%%%%%%%%%%%%%

\end{document}